\documentclass[nohyperref]{article}

\usepackage{microtype}
\usepackage{import}
\usepackage{cutwin}

\usepackage[utf8]{inputenc} 
\usepackage[T1]{fontenc}    
\usepackage{hyperref}       
\usepackage{url}            
\usepackage{amsfonts}       
\usepackage{nicefrac}       
\usepackage{xcolor}         
\usepackage{graphics}
\usepackage{listings}
\usepackage{algorithm}
\usepackage{algorithmic}
\usepackage{amsmath, amssymb, graphicx, amsthm}
\usepackage{subfigure}
\usepackage{booktabs} 
\usepackage{multirow}
\usepackage{caption}
\usepackage{tikz}
\usepackage{enumitem}
\usepackage{thm-restate}
\usepackage{mathtools}

\usepackage{bm}

\newcommand{\interior}[1]{%
  {\kern0pt#1}^{\mathrm{o}}%
}

\newcommand{\ms}{ \text{ms}}
\newcommand{\us}{ \mu\text{s}}
\newcommand{\ns}{ \text{ns}}

\DeclareFixedFont{\ttb}{T1}{txtt}{b}{n}{9} 
\DeclareFixedFont{\ttm}{T1}{txtt}{m}{n}{9}  
\usepackage{color}
\definecolor{deepblue}{rgb}{0,0,0.5}
\definecolor{deepred}{rgb}{0.6,0,0}
\definecolor{deepgreen}{rgb}{0,0.5,0}

\definecolor{light-gray}{gray}{0.95}
\definecolor{bole}{rgb}{0.35, 0.22, 0.21}
\definecolor{brown(web)}{rgb}{0.65, 0.16, 0.16}
\usepackage{listings}

\newcommand\pythonstyle{\lstset{
language=Python,
basicstyle=\ttm\color{bole},
morekeywords={self, torch, size, 
            Size, fc, x, y, tensor, f, nc, grad, FloatTensor, HalfTensor, DoubleTensor, MCModel },  
keywordstyle=\ttm\color{deepblue},
emph={MyClass,__init__, Two_Sum, Split, Two_Prod, 
      MCModule, MCOptim,  MCLinear, MCEmbedding, 
      add, subtract, multiply, divide, exp, 
      dot, mv, mm, matmul, 
      MCSGD, MCAdam,
      Two_Prod_fma, Renormalize, Simple_Renorm,  
      Grow_ExpN, ScalingN, Add_MCN, Div_MCN, ScalingN, DivN,
      DivN, Mult_MCN, Mult_MCN_Slow, Exp_MCN, 
      Square_MCN, Dot_MCN, MV_MCN, MM_MCN, BMM_MCN,
      Exp−MCN, Square−MCN, Grow−ExpN, Add−MCN, Div−MCN, Mul−MCN,
      4DMM_MCN,  AddMM_MCN,  Matmul_MCN, MC-Linear},          
emphstyle=\ttm\color{brown(web)},    
stringstyle=\color{deepgreen},
frame=tb,                         
showstringspaces=false,
backgroundcolor=\color{light-gray}
}}

\newcommand\pythonstylesmall{\lstset{
language=Python,
basicstyle=\fontsize{6}{8}\selectfont\ttm\color{bole},
morekeywords={self, torch, size, 
            Size, fc, x, y, tensor, backward, loss, f, nc, grad, FloatTensor, HalfTensor, DoubleTensor},  
keywordstyle=\ttm\color{deepblue},
emph={MyClass,__init__, Two_Sum, Split, Two_Prod, 
      MCModule, MCOptim,  MCLinear,Linear, MCEmbedding, MCSequential, ReLU, Softmax, GELU, 
      add, subtract, multiply, divide, exp, 
      dot, mv, mm, matmul, 
      MCSGD, MCAdam, SGD,
      Two_Prod_fma, Renormalize, Simple_Renorm,  
      Grow_ExpN, ScalingN, Add_MCN, Div_MCN,
      DivN, Mult_MCN, Mult_MCN_Slow, Exp_MCN, 
      Square_MCN, Dot_MCN, MV_MCN, MM_MCN, BMM_MCN,
      4DMM_MCN,  AddMM_MCN,  Matmul_MCN, MC-Linear},          
emphstyle=\ttm\color{brown(web)},    
stringstyle=\color{deepgreen},
frame=tb,                         
showstringspaces=false,
}}

\lstnewenvironment{python}[1][]
{
\pythonstyle
\lstset{#1}
}
{}

\newcommand\pythonexternal[2][]{{
\pythonstylesmall
\lstinputlisting[#1]{#2}}}
\newcommand\pythoninline[1]{{\pythonstyle\lstinline!#1!}}

\usepackage{wrapfig}

\usepackage[accepted]{icml2022}

\usepackage[capitalize,noabbrev]{cleveref}

\theoremstyle{plain}

\theoremstyle{definition}

\theoremstyle{remark}

\usepackage[textsize=tiny]{todonotes}

\icmltitlerunning{MCTensor: A High-Precision Deep Learning Library}

\begin{document}

\twocolumn[
\icmltitle{MCTensor: A High-Precision Deep Learning Library\\
with Multi-Component Floating-Point}



\icmlsetsymbol{equal}{*}

\begin{icmlauthorlist}
\icmlauthor{Tao Yu}{equal,cs}
\icmlauthor{Wentao Guo}{equal,cs}
\icmlauthor{Jianan Canal Li}{equal,cs,bee}
\icmlauthor{Tiancheng Yuan}{equal,se}
\icmlauthor{Christopher De Sa}{cs}
\end{icmlauthorlist}

\icmlaffiliation{cs}{Department of Computer Science,}
\icmlaffiliation{bee}{Biological Engineering,}
\icmlaffiliation{se}{and System Engineering, Cornell University}

\icmlcorrespondingauthor{Tao Yu}{tyu@cs.cornell.edu}
\icmlcorrespondingauthor{Wentao Guo}{wg247@cornell.edu}
\icmlcorrespondingauthor{Jianan Canal Li}{jl3789@cornell.edu}
\icmlcorrespondingauthor{Tiancheng Yuan}{ty373@cornell.edu}
\icmlkeywords{Machine Learning, ICML}

\vskip 0.3in
]



\printAffiliationsAndNotice{\icmlEqualContribution} 

\begin{abstract}



In this paper, we introduce MCTensor, a library based on PyTorch for providing general-purpose and high-precision arithmetic for DL training. MCTensor is used in the same way as PyTorch Tensor: we implement multiple basic, matrix-level computation operators and NN modules for MCTensor with identical PyTorch interface. Our algorithms achieve high precision computation and also benefits from heavily-optimized PyTorch floating-point arithmetic.
We evaluate MCTensor arithmetic against PyTorch native arithmetic for a series of tasks, where models using MCTensor in float16 would match or outperform the PyTorch model with float32 or float64 precision.

\end{abstract}

\section{Introduction}

High precision computations are of interest in many areas. For example, an emerging trend in studying dynamical systems is to use Taylor methods with high-precision arithmetic \cite{bailey2015high}, and delaunay triangulation in computational geometry \cite{schirra1998robustness}. Recently, high precision computations are even desired for some deep learning tasks, e.g. hyperbolic deep learning, so as to use hyperbolic space stably in practice \cite{yu2019numerically, yu2021representing}.

As of now, most high precision algorithms can be divided in two categories: (1) the standard multiple-digit ``BigFloat" format using a sequence of digits coupled with a single exponent term; (2) the multiple-component floating-point format (MCF) using an unevaluated sum of multiple ordinary floating-point numbers (e.g. float16, float32). Some examples of the first approach include the GNU Multiple Precision (GMP) Arithmetic Library \cite{Granlund12} and Julia's BigFloat type \cite{bezanson2017julia}. The idea of multiple-component float approach (also referred as ``expansion" in some literature), dates back to priest's works \cite{priest1991algorithms}, and some example implementations include multi-component library on C++ \cite{hida2007library} and MATLAB \cite{jiang2016implementation}. The multiple-digit approach can represent compactly a much larger range of numbers, whereas the multiple-component approach still adopts the limited precision floats to achieve high precision computations with error guarantees \cite{joldecs2015arithmetic}. 
However, the multiple-component approach has an advantage in speed over the multiple-digit approach, as it uses standard floats such as float16 and float32 and hence better takes advantage of existing floating-point accelerators \cite{hida2007library,shewchuk1997adaptive}.


In the emerging deep learning area, there is a demand for high precision computations in some deep learning applications (e.g. hyperbolic deep learning \cite{nickel2017poincare, liu2019hyperbolic}). Currently, there is \emph{no such deep learning library} for efficient high precision computations. Popular deep learning frameworks such as PyTorch \cite{paszke2019pytorch}, TensorFlow \cite{tensorflow2015-whitepaper} and Caffe \cite{jia2014caffe} only supports training with standard floats (e.g. float16, float32 and float64). Though high precision computations are enabled in Julia using multiple-digit ``BigFloat", deep learning libraries built on top of Julia such as Flux \cite{Flux.jl-2018,innes:2018}, MXNet.jl \cite{chen2015mxnet}, and TensorFlow.jl \cite{malmaud2018tensorflow} do not support training with BigFloat. Furthermore, BigFloat in Julia can \emph{only exist on CPUs}, but not on GPUs, which greatly limits its usages. 

In this work, we develop a multiple-component floating-point library, \emph{MCTensor}, for general-purpose, fast, and high-precision deep learning. We build MCTensor on top of PyTorch and it can be used in the same way as the PyTorch Tensor object. We hope MCTensor library can benefit applications in the following areas: (1) high precision training with low precision numbers. Training with low precision numbers is an emerging deep learning area on tasks like mobile computer vision \cite{tulloch2017high} and leveraging hardware accelerator like Google TPU \cite{jouppi2017datacenter}, where the deep learning model computes with low precision numbers such as float8 and float16. A large quantization error using low precision arithmetic would affect the convergence \cite{wu2018training} and may degrade the performance of the model; (2) numerical accurate and stable hyperbolic deep learning, where the non-Euclidean hyperbolic space is used in place of Euclidean space for many deep learning tasks and models due to its non-Euclidean properties. For example, graph embedding \cite{nickel2017poincare,nickel2018learning} and many developed hyperbolic networks, including hyperbolic neural networks \cite{ganea2018hyperbolic} and hyperbolic GCN \cite{chami2019hyperbolic,yu2022hyla}. However, the numerical error of computing with standard floating-point numbers in hyperbolic space is unbounded, even with float64, characterized as the ``NaN" problem \cite{sala2018representation,yu2019numerically}. A high precision computation in the hyperbolic space suggested using MCF \cite{yu2021representing} would be helpful. Our main contributions are as follows:
\begin{itemize}[nosep, leftmargin=*]
    \item We implement MCTensor in the same way as PyTorch Tensor with corresponding basic and matrix-level operations using MCF.
    \item We enable learning with MCTensor by developing the MCModule layers, MCOptimizers and etc with the same programming interface as PyTorch's counterparts.
    \item We demonstrate the performance of MCTensor for both high precision training with low precision numbers and on some hyperbolic tasks.  
\end{itemize}

\section{Methodologies}
Here we introduce some basics of our MCTensor library, built on top of PyTorch \cite{paszke2019pytorch} \footnote{Our PyTorch version is 1.11.0} that employs multi-component floating-point as its underlying tensor representation. Each MCTensor $x$ is represented as an expansion, an unevaluated sum of multiple tensors as follows:
\vspace{-2em}
\begin{equation} \label{eq:MCF}
    x = (x_0,x_1,\cdots,x_{nc-1}) = x_0 + x_1 + ... + x_{nc-1}  \vspace{-0.5em}
\end{equation}
where each $x_i$, as a component of $x$, can be a PyTorch floating-point Tensor in any precision, and $nc$ is the number of components for MCTensor $x$. It's required that all components to be ordered in a decreasing magnitude (with $x_0$ being the largest and $x_{nc-1}$ being the smallest). In this way, MCTensor allows roundoff error to be propagated to the later components and thus offers better precision compared to a standard PyTorch Tensor\footnote{unless otherwise specified, the PyTorch tensor, or ``Tensor", is referred to a PyTorch tensor with an arbitrary floating point data type in this paper}. 

We first implement basic operators \pythoninline{add}, \pythoninline{subtract}, \pythoninline{multiply}, \pythoninline{divide}, \ldots for MCTensor with MCF arithmetic and further vectorize them to matrix-level operators \pythoninline{dot, mm,} \ldots with same semantics as their PyTorch counterparts. These operators then allow us to implement higher-level \textbf{MCModule, MCOptim} as the counterpart for \pythoninline{torch.nn.Module} and \pythoninline{torch.optim} so that we can use them for any deep learning applications.

\subsection{MCTensor Object with Basic Operators}

\textbf{[MCTensor object]} A MCTensor $x$ can be abstracted as an object with specification \pythoninline{x\{fc, tensor, nc\}}. Specifically, \pythoninline{x.tensor} has $nc$ components of PyTorch tensors $x_0, x_1, \cdots , x_{nc-1}$ in the last dimension, and it has shape as $(*x_0.\text{shape}, nc)$. The $x.fc$ data term in MCTensor is a view of $x_0$, keeps track of the gradient for $x$, and if needed, serves as an approximate tensor representation of $x$.

\textbf{[Gradient]} Because a MCTensor is an unevaluated sum of Tensors, then the gradient of a function $f$ w.r.t. $x$ is $\frac{\partial f}{\partial x}  = \frac{\partial f}{\partial (x_0 + x_1 + \cdots + x_{nc-1})}= \frac{\partial f}{\partial x_i}$,
which is same as the the gradient of $f$ w.r.t. any component $x_i$.
So we only keep track of the gradient information of \pythoninline{x.fc} for a MCTensor \pythoninline{x}, which can then be computed naturally by PyTorch's auto-differentiation engine to get \pythoninline{x.fc.grad} as \pythoninline{x.grad}.

In order to develop more advanced operations of MCTensor, we develop two most basic functions first, \pythoninline{Two-Sum} and \pythoninline{Two-Prod}, whose inputs are two PyTorch Tensors and returns the result as a MCTensor with 2 components in the form of (result, error). 


\textbf{[Basic Operators]} We develop 
binary MCTensor operators to support input types (1) a MCTensor and a Tensor, (2) a Tensor and a MCTensor, and (3) a MCTensor and a MCTensor. For unary operators, we only accept a MCTensor as an input. The output for all these operators is a MCTensor with the same $nc$ as the input(s). We implement MCF algorithms for basic arithmetic, with the full list of their algorithm statements available in the appendix:
\begin{itemize}[nosep, leftmargin=*]
    \item \textbf{MCTensor:}  \pythoninline{Exp-MCN} (exp), \pythoninline{Square-MCN} (square)
    \item \textbf{MCTensor and Tensor:} \pythoninline{Grow-ExpN} (add), \pythoninline{ScalingN} (multiply), \pythoninline{DivN} (divide) 
    \item \textbf{MCTensor and MCTensor:} \pythoninline{Add-MCN} (add), \pythoninline{Div-MCN} (divide), \pythoninline{Mul-MCN} (multiply)
\end{itemize}

For example, the addition between a MCTensor and a Tensor, or \pythoninline{Grow-ExpN} (Grow Expansion with Normalization), is given in Alg.~\ref{alg:growexpn_main}.

\begin{algorithm}[H]
   \caption{\textbf{Grow-ExpN}}
   \label{alg:growexpn_main} 
\begin{algorithmic}
   \STATE {\bfseries Input:}  $nc$-MCTensor $x$, PyTorch Tensor $v$
   \STATE \textbf{initialize} $Q\leftarrow v$
   \FOR{$i=1$ to $nc$}
    \STATE $k\leftarrow nc+1-i$
    \STATE $(Q, h_k) \leftarrow \textbf{Two-Sum}(x_{k-1}, Q)$
   \ENDFOR
   \STATE $h  \leftarrow (Q,h_1,\cdots,h_{nc})$
   \STATE {\bfseries Return:} $\textbf{Simple-Renorm}(h, nc)$  
\end{algorithmic}

\end{algorithm}

\pythoninline{Grow-ExpN} takes a $nc$-component MCTensor $x$ and a normal Tensor $v$ as input. The approximated result $Q$ of this addition, is initialized to $v$. The Grow Expansion happens first in the for loop, starting with the last component of $x$, $x_{nc-1}$, to the first component $x_0$. The algorithm \pythoninline{Two-Sum} sums $Q$ and a component $x_{k-1}$, resulting in the updated approximation $Q$ and the error term $h_k$. The resulting $h$ is therefore grown into $nc+1$ components, which are naturally ordered in a decreasing manner, except that there could be some intermediate zero components. Hence, in order to meet the requirements of a MCTensor, we use \pythoninline{Simple-Renorm} algorithm to move zeros backwards and output a $nc$ components MCTensor.

We also implement two different algorithms for \pythoninline{Mul-MCN}: a fast version that uses two \pythoninline{Div-MCN} operations to perform multiplication, and a slower version that have better error bounds. However, in practice we see little difference between them, so we run all our experiments with the fast version. Details of these basic MCF algorithms are provided in the appendix.

\begin{figure}[t]
\centering
\includegraphics[width=0.45\textwidth]{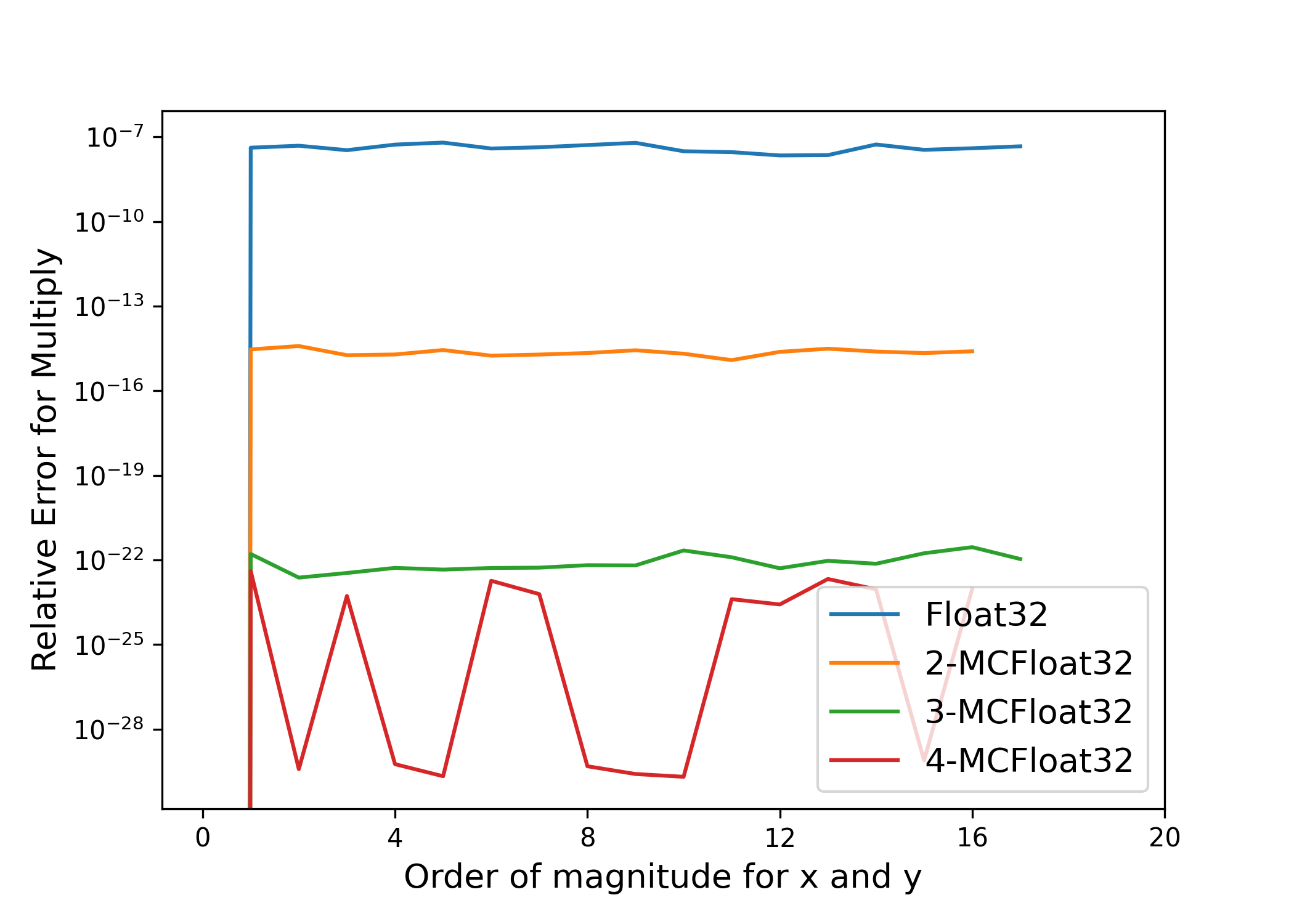} 
\vspace{-1em}
\caption{Relative Errors for Multiplication \pythoninline{Mul-MCN(x,y)}, Compared with High Precision Julia BigFloat (3000 bits precision). Order of magnitudes for \pythoninline{x} and \pythoninline{y} are the same.} 
\label{fig:mult-err}
\vspace{-2em}
\end{figure}

To demonstrate how a MCTensor's precision increases with the number of components $nc$, we plot the relative numerical errors for MCTensor with different $nc$ and PyTorch Tensor. We set the data type for both MCTensors and Tensors to be Float32, and compute the errors w.r.t. the results derived by high precision Julia BigFloat with number of bits of the significand set to 3000. As can be seen from Figure~\ref{fig:mult-err}, even with $nc=2$, the relative error is orders of magnitude smaller than PyTorch Tensors. 

\subsection{MCTensor Matrix Operators}

After defining the basic MCF arithmetic, we are able to implement commonly used matrix level operators for MCTensor including \pythoninline{AddMM-MCN} (torch.addmm), \pythoninline{Dot-MCN} (torch.dot), \pythoninline{MV-MCN} (torch.mv), \pythoninline{MM-MCN} (torch.mm), \pythoninline{BMM-MCN} (torch.bmm) and \pythoninline{Matmul-MCN} (torch.matmul). Details of them are provided in the appendix.


For matrix operators, we leverage the broadcastability and vectorization from native PyTorch operations embedded in our basic MCF operators across each $nc$, and then apply sequential error propagation. For example, for \pythoninline{MV-MCN} with input as $x$ and $v$, we first use \pythoninline{ScalingN} to compute the broadcasted product of $x$ and $v$ for each $nc$, and then we sequentially sum up the results and propagate errors with the \pythoninline{Add-MCN} operator. In this way, we can make our multiplication part (\pythoninline{ScalingN}) independent of the input size and only employ for-loop for addition part (\pythoninline{Add-MCN}) since error propagation in the addition is sequential by the algorithm. Theoretically, accurate addition of $N$ MCTensors is of order at least $O(nc\cdot\log N)$. 

MCTensor operators will be slower because of the need to propagate errors in computation, and have more memory burden than PyTorch operators because of the nature of MCF representation. In Table~\ref{tab:matrixop}, we can see a tradeoff between program speed
and precision. However, we would like to note that there is still much space to optimize these algorithms for better timing in practice. This work aims to provide ML community with the possibility to do high-precision computations for learning with MCTensor over GPUs. More details can be found in \ref{mm}.

\begin{figure*}[htbp]
\centering
{\pythonexternal{code/MCLinear.py}} \vspace{-1em}
\caption{An Example for Implementing MCLinear }
\label{fig:code_mclinear}
\end{figure*}

\begin{table*}[htbp]
 \centering
    \scalebox{0.9}{\small\begin{tabular}{lccccl} 
    \toprule
     Operators & Inputs sizes & FloatTensor & 1-MCTensor & 2-MCTensor & 3-MCTensor  \\
    \midrule
    \textbf{Dot-MCN} & $5000, 5000$ & $ 1.61\us\pm3.29\ns$ 
    & $ 442\us\pm5.61\us$
    & $ 656\us\pm1.16\us$
    & $ 858\us\pm12.2\us$\\                
\textbf{MV-MCN} & $(5000\times 500),500$    
& $ 157\us\pm4.32\us$
& $ 320\ms\pm5.78\ms$
& $ 460\ms\pm10.7\ms$
& $ 580\ms\pm12.1\ms$ \\
    \textbf{Matmul-MCN} & ( 500$\times$ 200), ( 200$\times$ 50)    
    & $ 97.3\us\pm1.1\us$
    & $ 495\ms\pm10.8\ms$
    & $ 735\ms\pm21.7\ms$
    & $ 934\ms\pm28\ms$ \\
    \bottomrule
    \end{tabular}}
    \caption{\centering  MCTensor Matrix Operators Running Time (mean $\pm$ sd) }
    \label{tab:matrixop}
    \vspace{-1em}
\end{table*}

\subsection{MCModule, MCActivation, and MCOptim}

We enable learning with MCTensor by developing neural network basic modules (MCModule), activation functions (MCActivation) and optimizers (MCOptim), in the exact programming interfaces as their PyTorch counterparts in (\pythoninline{torch.nn.Module}, \pythoninline{torch.nn.functional} and \pythoninline{torch.optim}).
The semantics are identical to that of PyTorch's module and optimizer, except that we are using MCTensor arithmetic. Specifically, we give some examples:
\begin{itemize}[nosep, leftmargin=*]
    \item \textbf{MCModule:} {\pythoninline{MCLinear, MCEmbedding, MCSequential}}
    \item \textbf{MCOptim:} {\pythoninline{MC-SGD, MCAdam}}
    \item \textbf{MCActivation:} {\pythoninline{MCSoftmax, MCReLU, MC-GELU}} 
\end{itemize}
Here Fig.\ref{fig:code_mclinear} is the MCTensor implementation of linear layer, where weights and biases are represented by MCTensors. Similar to its PyTorch peer, a \pythoninline{MCLinear} takes input of size of input shape, in\_features, size of output shape, out\_features, learn with or without biases (boolean: bias) and number of component $nc$ for setting up the underlying MCTensors.  To handle the operations within MCLinear layer, we override PyTorch's \pythoninline{nn.functional.linear} to perform matrix level multiplication (Algo. \ref{alg:matmulmcn}) between a MCTensor weight and the Tensor input, then the addition of the product with bias (Algo.\ref{alg:addmcn}) if needed. The output of the MCLinear layer is a MCTensor with the same $nc$.

\par
Since we implemented MCModule, MCActivation, and MCOptimizer to follow the same specification as their PyTorch counterparts, building a MCTensor model is the same way as one would build a PyTorch module. The only difference to the users is the need to specify the number of components $nc$. An demonstration of our MCTensor model optimization programming paradigm can be found in Fig. \ref{fig:code_mc_opt}. \vspace{-1em}
\begin{figure} [h]
\centering
\pythonexternal{code/MCOptimization.py} \vspace{-1em}
\caption{\centering MCTensor model optimization programming paradigm}
\label{fig:code_mc_opt}
\vspace{-2em}
\end{figure}

\subsection{Error Analysis} \label{subsec:error}
Principally, one can achieve arbitrary precision using MCTensor by simply increasing the number of components, however, note that each component of a MCTensor is a standard float, which has a natural range. Take Float16 for example, the minimum representable strictly positive value is $2^{-24}\approx 5.96 \times 10^{-8}$, i.e., the smallest error that can be captured by MCFloat16 is $2^{-24}$. Hence in practice, 2-MCFloat16 is usually sufficient and similar performances are observed when more components of MCFloat16 were adopted. While simply adding an appropriate scale factor $2^{-k}$ (depending on the precision requirements) to smaller components can help capture even smaller errors, it is out of scope of this paper. 

To validate improved precision of MCTensor models, we consider the linear regression task since it is possible to obtain loss arbitrarily close to zero.
We use a single \pythoninline{MCLinear} without bias term and the Mean Squared Error (MSE) loss on fully observed synthetic data: $\mathcal{L}(W) = \text{MSELoss}(y, XW^T)$,
where $X$ is a $(10000 \times 2)$ matrix with each entries sampled from $x  \sim \mathcal{N}(-0.5, 0.5^2)$, and $y$ is calculated from $y=X{W^*}^T$ where $W^*$ is the $(10000\times1)$ target weight sampled from $w^* \sim \mathcal{N}(-0.5, 0.5^2)$. We use gradient descent with $lr=0.05$ for optimization.

\begin{figure} 
\vspace{-1em}
\centering
 \includegraphics[width=0.45\textwidth]{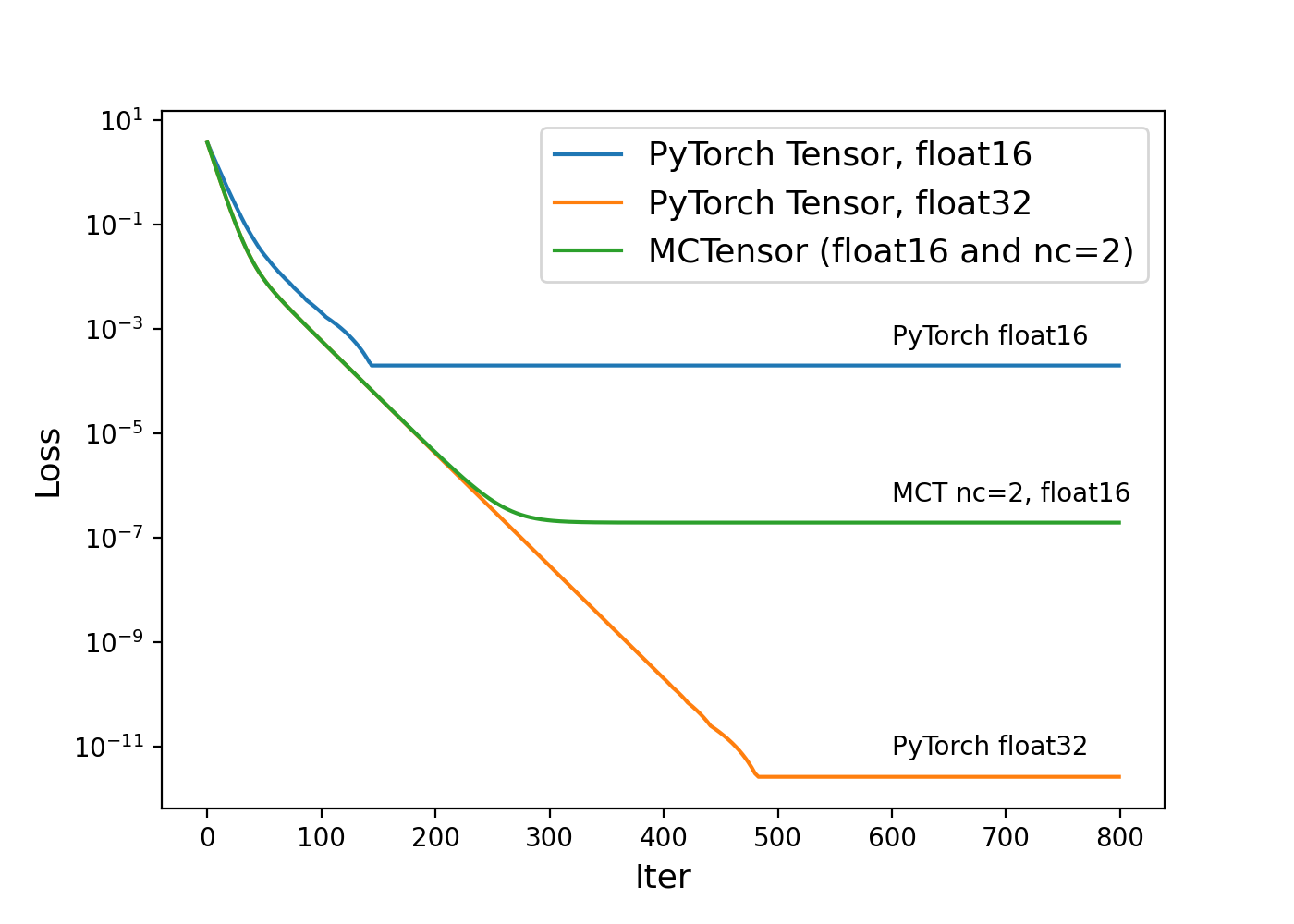}
\vspace{-1em}
\caption{\centering Loss Curves on Linear Regression Task.}
\label{linlossfig}
\vspace{-1em}
\end{figure}

In Figure~\ref{linlossfig}, we plot the training loss curves for the model with the same structure and initialization, but with MCTensor or Tensor as data structure. The comprehensive results can be seen in Table \ref{tab:linreg}. The final train loss for 2-MCFloat16 is orders of magnitudes smaller than Float16.


Since by just using 2-MCFloat16, we can achieve much better precision than Float16 Tensor (HalfTensor), we run all following experiments in Float16 with $nc=$2 or 3 expect for Hyperbolic MCEmbedding. 







\section{Experiments}

A MCTensor is able to achieve improved precision compared with a PyTorch Tensor with the same data type. If higher precision implies better results under the same experiment settings, we would expect our $n$-MCTensor ($n \geq 2$) model can achieve better performance than a native PyTorch model with same precision. We demonstrate the increasing precision under the same data type for MCTensor by carrying out experiments in two settings: (1) high precision learning with low precision numbers, this include logistic regression model
and multi-layer perceptrons (MLP) in Section~\ref{subsec:mlp}; and (2) high precision hyperbolic embedding reconstruction task in Section~\ref{subsec:hyperbolic}.
For all these experiments, MCTensor models use the same initialization as the Tensor models, and use \pythoninline{MCModule} as its PyTorch's \pythoninline{nn.Module} counterpart, with MCTensors as layer weights. Our code and models are open-source and available at github \footnote{\url{https://github.com/ydtydr/MCTensor}}.




\subsection{High-precision Computations with Low Precision}
Low precision machine learning employs models that compute with low precision numbers (e.g. float8 and float16), which become popular in many edge-applications \cite{hubara2017quantized,zhang2019qpytorch}. However, the quantization error inherent with low precision arithmetic could affect the convergence and performance of the model \cite{wu2018training,de2018high}. As a matter of fact, most current low precision learning frameworks including \cite{courbariaux2014training,de2018high} adopt high precision numbers during gradient computations and optimizations, but pursue a better way to convert the results to low-precision numbers with less quantization errors. In comparison, with MCTensor, one can get \emph{accurate} update with purely low-precision numbers thoroughly without even touching high precision arithmetics. This is particularly helpful on devices where only low-precision arithmetics are supported.


\paragraph{Logistic Regression.}
\label{subsec:logistic}
\begin{figure}
    \centering
    \includegraphics[width=0.4\textwidth]{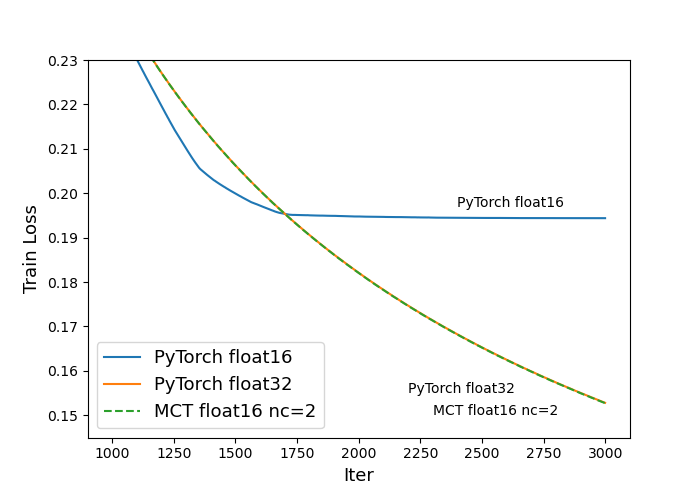} 
    \vspace{-1em}
    \caption{\centering Training Losses for Logistic Regression Models on the Breast Cancer Dataset}
    \label{fig:my_label}
    \vspace{-1em}
\end{figure}

We conduct a logistic regression task on a synthetic dataset and the cancer dataset, both are datasets with binary labels. The synthetic dataset consists of $1,000$ data points, where each data point contains two features. This dataset is constructed through the \pythoninline{make\_classification} \cite{guyon2003design} function from scikit-learn package with both \pythoninline{n\_informative} and \pythoninline{n\_clusters\_per\_class} are set to 1.  The breast cancer dataset \cite{breast_cancer} consists of 569 data points and each data point contains 30 features. More details can be found in \ref{logreg}.

\paragraph{Multi-layer Perceptron.}
\label{subsec:mlp}
We also construct a Multi-layer Perceptron using MCTensor (MC-MLP) and evaluate it on classification tasks on the breast cancer dataset and a reduced MNIST, which has only 10,000 data points in total, with 1,000 images sampled randomly per class \cite{deng2012mnist}. The MC-MLP consists of three \pythoninline{MCLinear} layers with model weight in MCFloat16. After each \pythoninline{MCLinear} layer, the resulting MCTensor is transformed into a normal tensor, passed it to activation function and fed to the next layer.
\begin{figure}[t]
    \centering
    \includegraphics[width=0.4\textwidth]{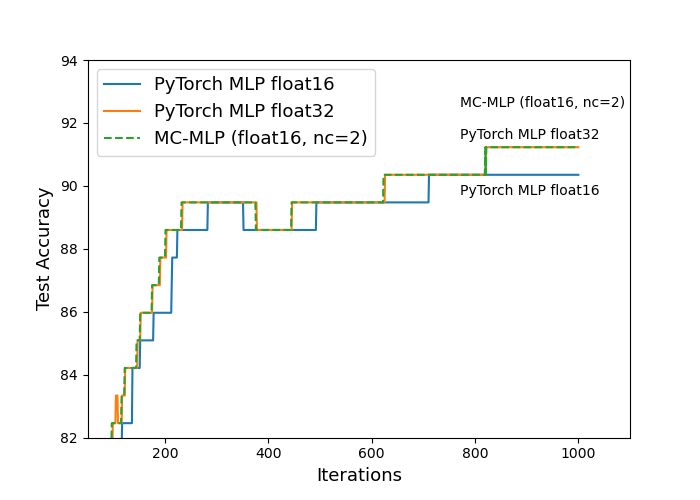}
    \vspace{-1.5em}
    \caption{Test Accuracy for MC-MLP on Breast Cancer Dataset. Notice that the curve for MC-MLP model with nc=2, float16 essentially overlaps with the curve for PyTorch MLP model with float32.}
    \label{fig:my_label}
\end{figure}

\begin{figure}[t]
    \centering
    \includegraphics[width=0.4\textwidth]{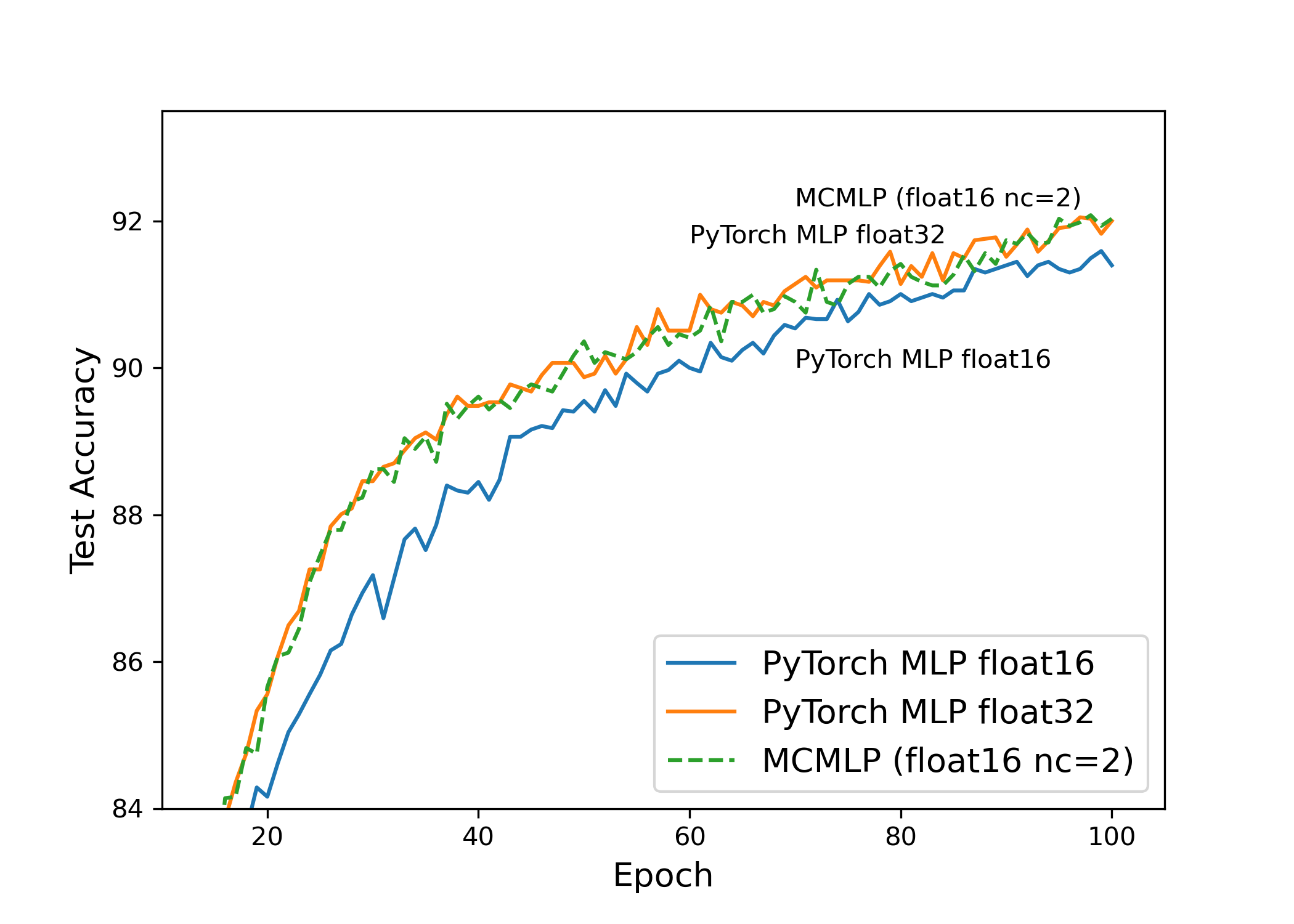}
    \vspace{-1.5em}
    \caption{Test Accuracy for MC-MLP on Reduced MNIST Dataset. Notice that the curve for MC-MLP model with nc=2, float16 essentially overlaps with the curve for PyTorch MLP model with float32.}
    \label{fig:mcmlp-test-mnist}
\end{figure}

We experiment on both the Breast Cancer dataset and the Reduced MNIST dataset, and as demonstrated in Table~\ref{tab:mlp-b} and Figure~\ref{fig:mcmlp-test-mnist}, in both experiments, MC-MLP outperforms PyTorch float16 models, and arrive at a lower training loss and a higher test accuracy (same as PyTorch float32 and float64 models). Notice that for both datasets, after $nc$ exceeds certain value ($nc=2$), adding extra more $nc$ would not lead to further improvement. 
More details can be found in Section \ref{MC-MLP}.

\begin{table}[t]
 \centering
    
    \scalebox{1.0}{
        \small\begin{tabular}{lccr}
        \toprule
         Model & Training Loss &  Testing accuracy   \\
        \midrule
         {MLP Float16} & 0.144 & 90.35\\
        \textbf{MLP Float32} & \textbf{0.124} & \textbf{91.23}\\
        \textbf{MLP Float64} & \textbf{0.124} & \textbf{91.23}\\
        \midrule
         {MC-MLP (nc=1)} & 0.144 & 90.35\\
        \textbf{MC-MLP (nc=2)} & \textbf{0.124} & \textbf{91.23}\\
        \textbf{MC-MLP (nc=3)} & \textbf{0.124} & \textbf{91.23}\\
        \bottomrule
        \end{tabular}
    }
    \caption{\centering MC-MLP Models on Breast Cancer Dataset with MCSGD}
    \label{tab:mlp-b}
     \vspace{-2em}
\end{table}

\subsection{Hyperbolic Embedding} 
\label{subsec:hyperbolic}
Another use case that benefits from MCTensor is hyperbolic deep learning, where the non-Euclidean hyperbolic space is adopted in place of Euclidean space for various purposes. It has been shown in \cite{nickel2017poincare,chami2019hyperbolic,yu2022hyla} that hyperbolic space is better suited for processing hierarchical data (e.g. trees, acyclic graphs). However, representing hyperbolic space with ordinary floating-points (even float64) leads to unbounded representation error, particularly when points get far away from the origin, known as the ``NaN" problem \cite{yu2019numerically} that not-a-number error occurs during computations. With MCTensor, we can achieve high-precision computations and avoid the ``NaN" problem by simply using more components. 

Following \citet{yu2021representing}, we conduct hyperbolic embedding experiment on the WordNet Mammals dataset with 1181 nodes and 6541 edges. We use Poincaré upper-half space (\textbf{Halfspace}) model to embed its transitive closure for reconstruction. The Poincaré Halfspace model is the manifold $(\mathcal{U}^n, g_u)$, where $\mathcal{U}^n$ is the upper half space of the $n$-dimensional Euclidean space, with the metric tensor and distance function being:
\[
\textstyle
g_u(\mathbf{x}) = \frac{g^E}{x_n^2},~~~~ d_u(\mathbf{x},\mathbf{y}) = \text{arcosh}
\left(1+ \frac{||\mathbf{x}-\mathbf{y}||^2}{2x_ny_n} \right),
\]
where $g^E$ is the Euclidean metric tensor. For the observed edges $\mathcal{D} = \{(\mathbf{x},\mathbf{y})\}$, we learn the embeddings $\Theta$ for all nodes, subject to minimizing the following loss function
\[
\textstyle
\mathcal{L}(\Theta) = \sum_{(\mathbf{x},\mathbf{y})\in D} \log \frac{e^{-d_u(\mathbf{x},\mathbf{y})}}{\sum_{\mathbf{y'}\in \mathcal{N}(\mathbf{x})} e^{-d_u(\mathbf{x},\mathbf{y'})}},
\]
where $\mathcal{N}(\mathbf{x})$ are randomly chosen 50 negative examples in addition to the positive example $(\mathbf{x},\mathbf{y})$. We report the results in Table~\ref{tab:mammals}, where MAP is the mean averaged precision and MR is the mean rank metric.


\begin{table}[ht]

 \centering
   
    \scalebox{0.9}{\small\begin{tabular}{lcr} 
    \toprule
     Model & MAP (mean $\pm$ sd) & MR (mean $\pm$ sd)   \\
    \midrule
\textbf{Halfspace (f32)} & $ 91.91 \% \pm 0.64 \% $  & $ 1.399  \pm  0.04 $\\
\textbf{Halfspace (f64)} & $ 92.79 \% \pm 0.41 \% $  & $ 1.340  \pm  0.07 $  \\     
\midrule
\textbf{MC-Halfspace (f64 nc=1)} & $ 93.02 \% \pm 0.40 \% $  & $ 1.296 \pm  0.02 $\\  
\textbf{MC-Halfspace (f64 nc=2)} & $ 92.77 \% \pm 0.28 \% $  & $ 1.304 \pm  0.02 $  \\
\textbf{MC-Halfspace (f64 nc=3)} & $ \textbf{93.31} \% \pm 0.75 \% $  & $ \textbf{1.282}  \pm 0.03 $   \\
    \bottomrule
    \end{tabular}}
    \vspace{-1em}
     \caption{\centering Performance of Hyperbolic Models}
    \label{tab:mammals}
\end{table}

\section{Conclusion}
We introduce MCTensor based on PyTorch to achieve high-precision arithmetic while leveraging the benefits of heavily-optimized floating-point arithmetic. We verify its capability of high precision computations using low precision numbers and relieving the NaN problem in hyperbolic space. We hope this library could interest researchers to use MCTensor for ML applications requiring high-precision computations.

A promising future work is to design and optimize MCF algorithms to granularize the tradeoff between efficiency and precision to make MCTensor competent even for less-noisy and general tasks.
We hope this library can address the need for fast high-precision library absent for DL community and prompt DL practitioners to rethink the concept of high-precision.

\section*{Acknowledgement}
This work is supported by NSF IIS-2008102.

\nocite{langley00}
\bibliography{example_paper}
\bibliographystyle{icml2022}

\newpage
\appendix
\onecolumn
\section{Appendix}
\subsection{Julia BigFloat error}
We further conduct experiments to evaluate the numerical errors of basic MCTensor arithmetic. Specifically, we first compute \pythoninline{Add-MCN(x,y)} of two random MCTensor \pythoninline{x,y} of roughly the same magnitude $m$.  \pythoninline{x,y} are sampled by first sampling two random Julia BigFloat numbers with high precision (e.g. 3000 precision) using the equation $(10-\mathcal{N}(0,1))^{m}$, then converted to their corresponding MCTensors in Float32. In order to get the \emph{exact} numerical error, we transform the MCTensor result to a Julia BigFloat number, then compute the relative error of it to the high precision addition of \pythoninline{x,y} (and not the addition of the two BigFloat numbers initially sampled) in Julia. In the same way, we compute the numerical errors of \pythoninline{Mult-MCN(x,y)} and \pythoninline{ScalingN(x,y)} with \pythoninline{y} being MCTensor in the former case and standard (PyTorch) tensor in the later case. \pythoninline{x,y} are sampled in the same way throughout the three cases. 

\begin{figure}[H]
\centering
\includegraphics[width=0.33\textwidth]{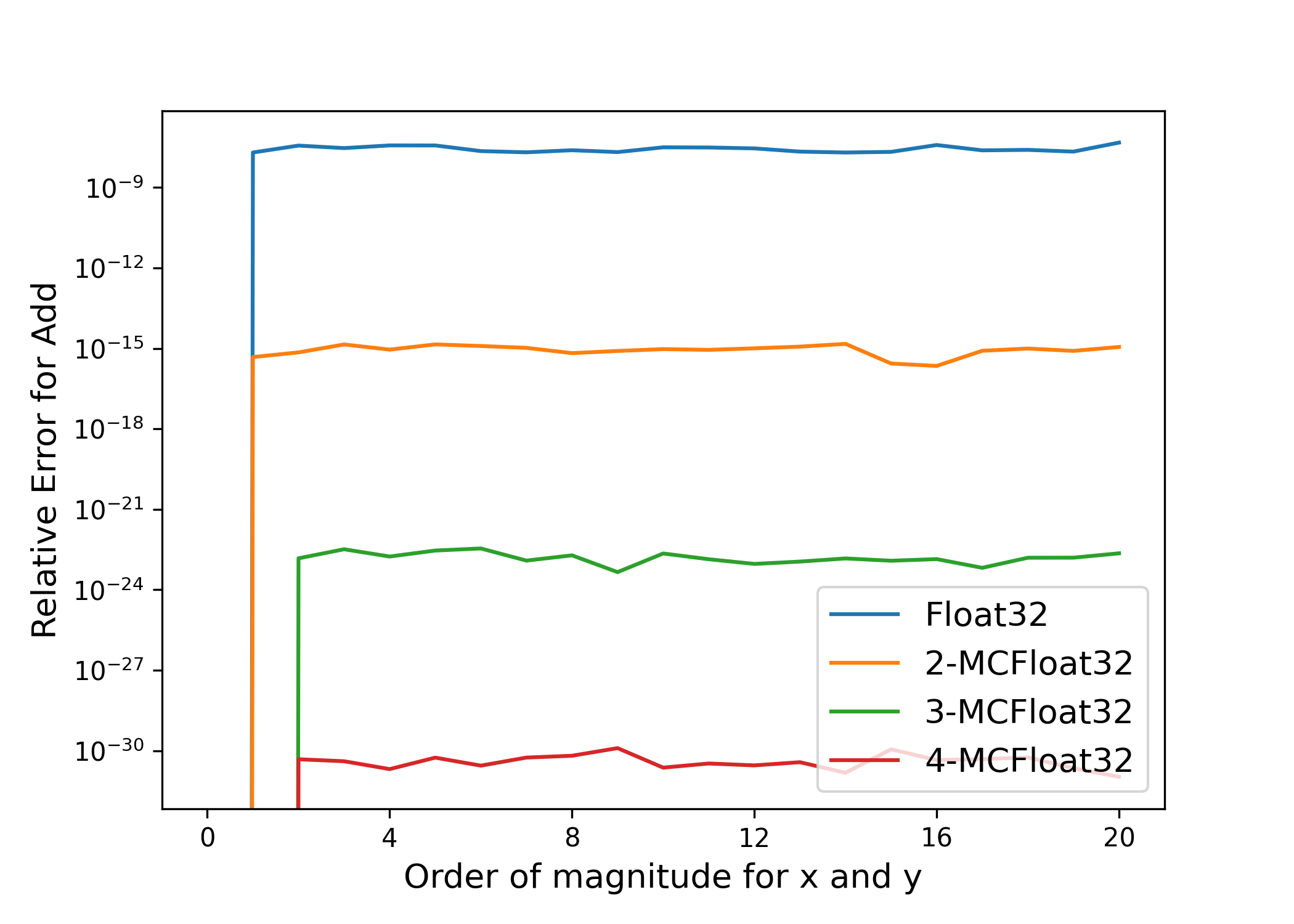}~ 
\includegraphics[width=0.33\textwidth]{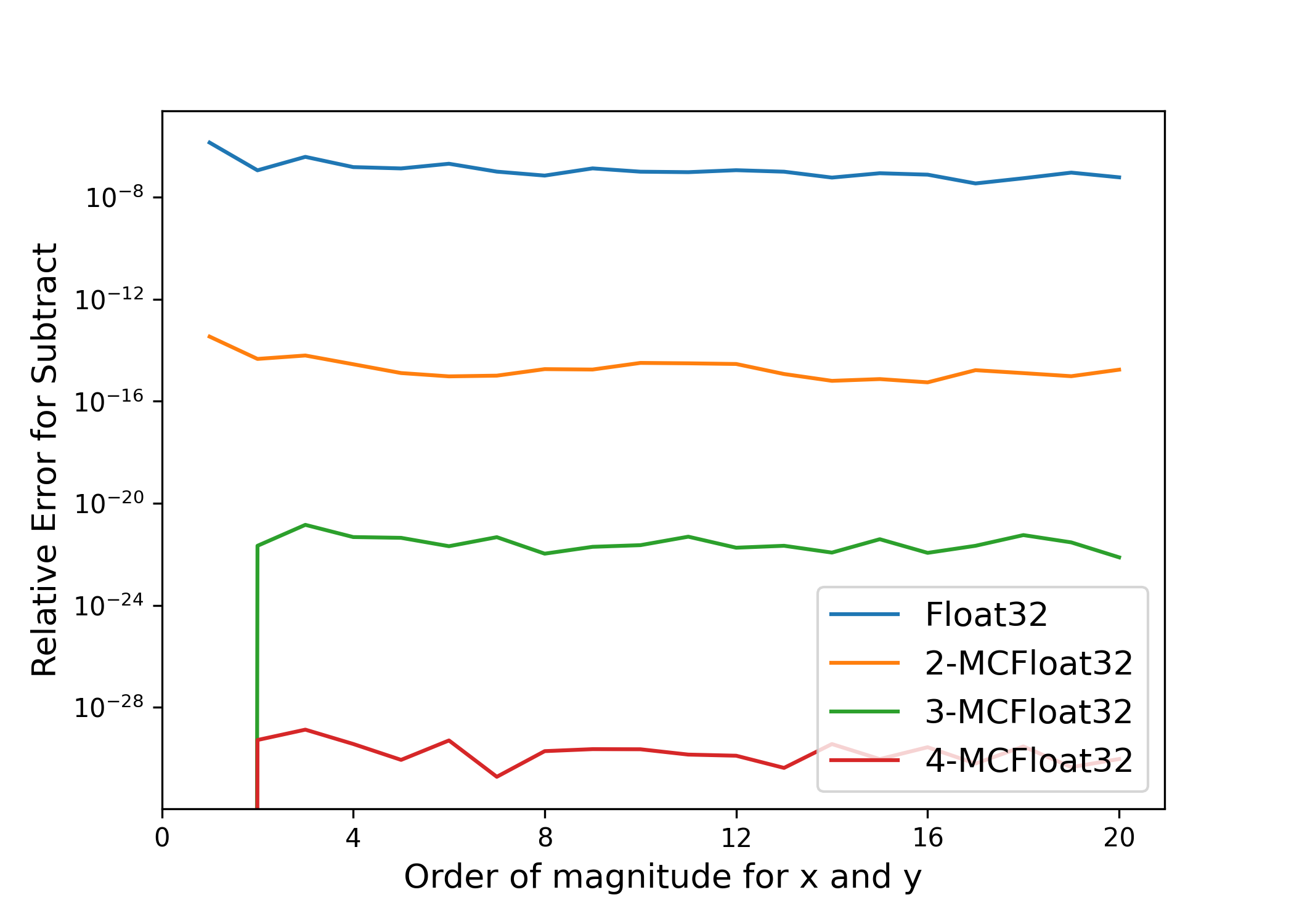}~
\includegraphics[width=0.33\textwidth]{figs/julia/same/mult.png}~
\vspace{-1em}
\caption{Relative Error of MCTensor arithmetic with different number of components, compared with high precision Julia BigFloat results (i.e. 3000 precision). Left: \pythoninline{Add-MCN(x,y)}, middle: \pythoninline{ScalingN(x,y)} and right: \pythoninline{Mult-MCN(x,y)}. Order of magnitudes for \pythoninline{x} and \pythoninline{y} are kept the same.}
\end{figure} 

For a thorough comparison, we also derive below the same numerical errors when the order of magnitudes for \pythoninline{x} varies and order of magnitudes for \pythoninline{y} is kept at 2.

\begin{figure}[H]
\centering
\includegraphics[width=0.33\textwidth]{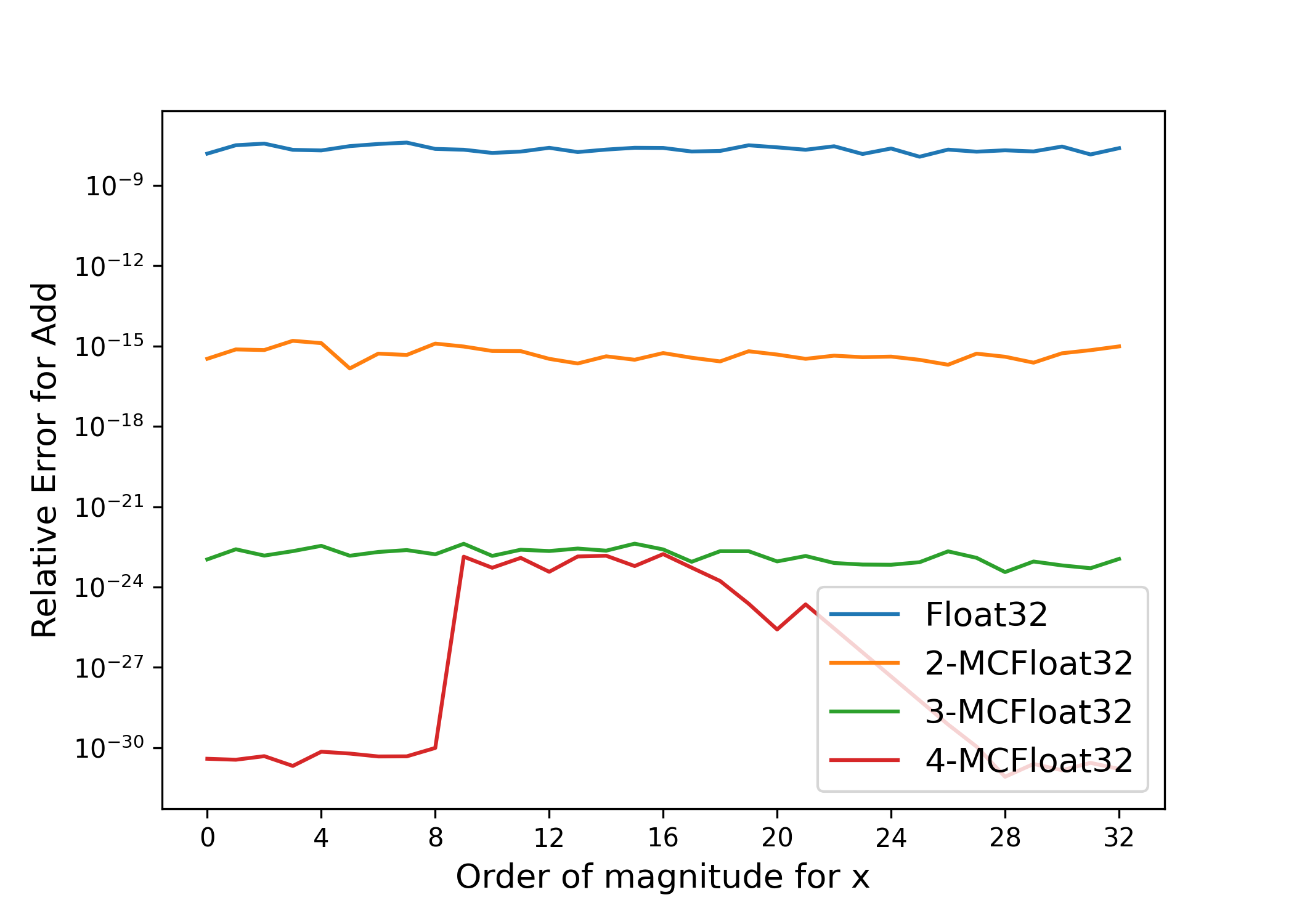}~ 
\includegraphics[width=0.33\textwidth]{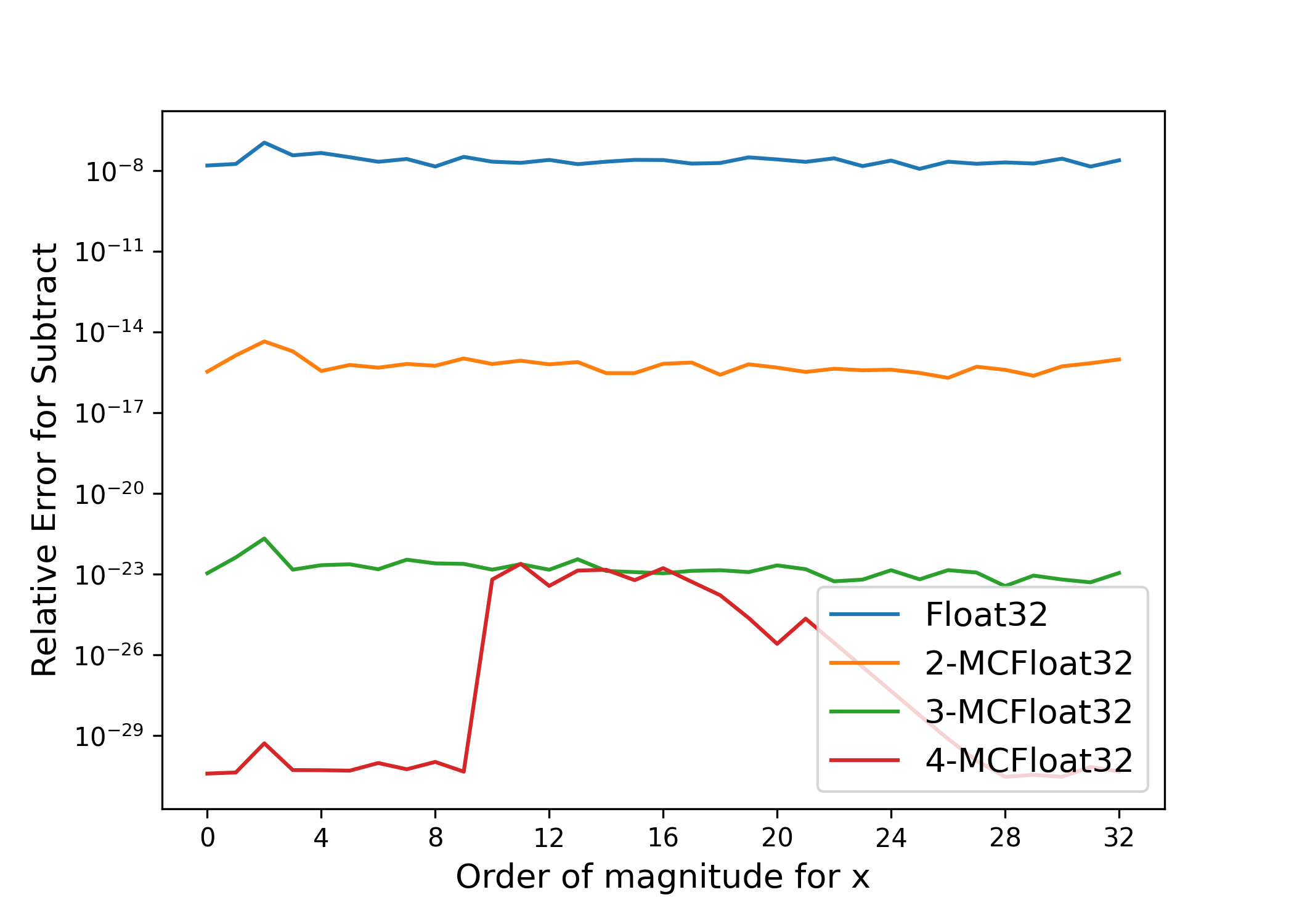}~
\includegraphics[width=0.33\textwidth]{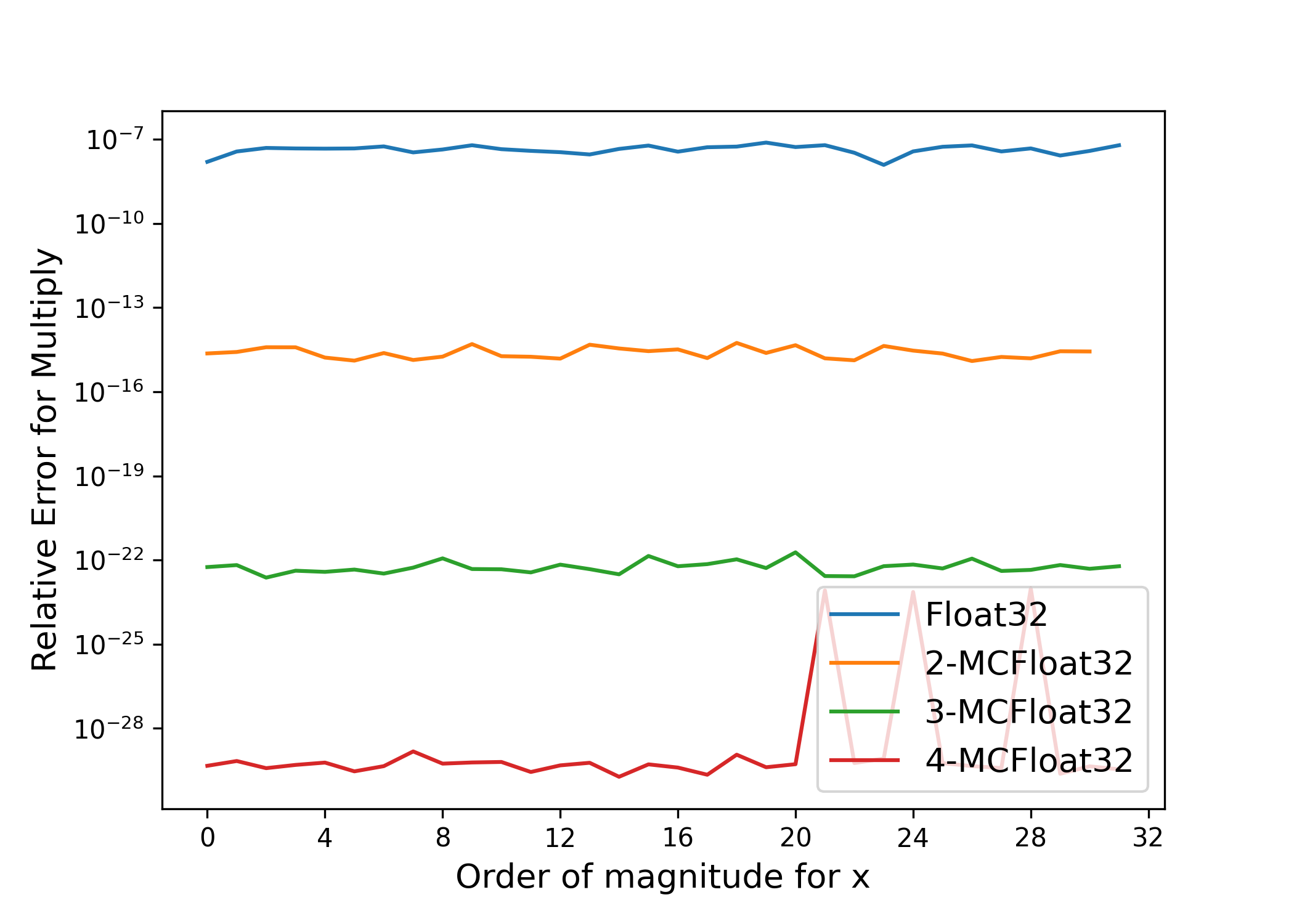}~
\vspace{-1em}
\caption{
Relative Error of MCTensor arithmetic with different number of components, compared with high precision Julia BigFloat results (i.e. 3000 precision). Left: \pythoninline{Add-MCN(x,y)}, middle: \pythoninline{ScalingN(x,y)} and right: \pythoninline{Mult-MCN(x,y)}. Order of magnitudes for \pythoninline{x} varies and order of magnitudes for \pythoninline{y} is kept at 2.}
\end{figure}


\newpage
\twocolumn
\subsection{MCTensor Operators}
\subsubsection{Basic Operators}
The input of \pythoninline{Two-Sum} is two PyTorch Tensors with same precision, $a$ and $b$. Algorithm~\ref{alg:twosum} returns the sum $s = \text{fl}(a + b)$ and the error, $err(a + b)$.
\vspace{-3mm}
\begin{algorithm}[H]
        \caption{\textbf{Two-Sum}}
        \label{alg:twosum}
        \begin{algorithmic}
          \STATE {\bfseries Input:} PyTorch Tensors $a,b$ 
           \STATE $x\leftarrow a+b$
           \STATE $b_{\text{virtual}}\leftarrow \textbf{fl}(x-a)$
           \STATE $a_{\text{virtual}}\leftarrow \textbf{fl}(x-b_{\text{virtual}})$
          \STATE $b_{\text{roundoff}}\leftarrow \textbf{fl}(b-b_{\text{virtual}})$
          \STATE $a_{\text{roundoff}}\leftarrow \textbf{fl}(a-a_{\text{virtual}})$
          \STATE $y\leftarrow \textbf{fl}(a_{\text{roundoff}}+b_{\text{roundoff}})$
           \STATE {\bfseries Return:} $(x,y)$
        \end{algorithmic}
\end{algorithm}
\vspace{-3mm}

The \pythoninline{Split} Algorithm~\ref{alg:split} takes a standard PyTorch floating point value with $p$-bit significand and splits it into its high and low parts, both with $\frac{p}{2}$-bit of significand.
\vspace{-3mm}
\begin{algorithm}[H]
        \caption{\textbf{Split}}
        \label{alg:split}
        \begin{algorithmic}
          \STATE {\bfseries Input:} PyTorch Tensor $a$
          \IF{$a.dtype$ is HalfTensor (float16)} 
          \STATE $constant\leftarrow 6$
          \ELSIF{$a.dtype$ is FloatTensor (float32)}
          \STATE $constant\leftarrow 12$
          \ELSIF{$a.dtype$ is DoubleTensor (float64)}
          \STATE $constant\leftarrow 26$
          \ENDIF
          \STATE $t\leftarrow\textbf{fl}(2^{constant} + 1)\cdot a$
          \STATE $a_{hi}\leftarrow \textbf{fl}(t- \textbf{fl}(t-a))$
          \STATE $a_{lo}\leftarrow \textbf{fl}(a - a_{hi})$
           \STATE {\bfseries Return:} $(a_{hi}, a_{lo})$
        \end{algorithmic}
\end{algorithm}
\vspace{-3mm}

Based on \pythoninline{Split}, the following Algorithm~\ref{alg:twoprod} computes and returns $p = \textbf{fl}(a \times b)$ and $e = \text{err}(a \times b)$.
\vspace{-3mm}
\begin{algorithm}[H]
        \caption{\textbf{Two-Prod}}
        \label{alg:twoprod}
        \begin{algorithmic}
          \STATE {\bfseries Input:} PyTorch Tensors $a$, $b$
          \STATE $p\leftarrow \textbf{fl}(a \cdot b)$
          \STATE $(a_{hi}, a_{lo})\leftarrow\textbf{Split}(a)$
          \STATE $(b_{hi}, b_{lo})\leftarrow\textbf{Split}(b)$
          \STATE $err_1\leftarrow \textbf{fl}(p -  \textbf{fl}(a_{hi}\cdot b_{hi}))$
          \STATE $err_2\leftarrow \textbf{fl}(err_1- \textbf{fl}(a_{lo}\cdot b_{hi}))$
          \STATE $err_3\leftarrow \textbf{fl}(err_2- \textbf{fl}(a_{hi}\cdot b_{lo}))$
          \STATE $e\leftarrow \textbf{fl}(\textbf{fl}(a_{lo}\cdot b_{lo})-err_3)$
           \STATE {\bfseries Return:} $(p,e)$
        \end{algorithmic}
\end{algorithm}
\vspace{-3mm}

Fuse multiply-add, or FMA, is a floating-point operation that performs multiplication and addition in one step.  With proper hardware, this Algorithm~\ref{alg:twoprodfma} can speed up \pythoninline{TwoProd}.
\vspace{-3mm}
\begin{algorithm}[H]
        \caption{\textbf{Two-Prod-fma}}
        \label{alg:twoprodfma}
        \begin{algorithmic}
          \STATE {\bfseries Input:} PyTorch Tensors $a$, $b$ 
          \STATE {\bfseries Requires:} Machine supports FMA instructions set
          \STATE $p\leftarrow \textbf{fl}(a \cdot b)$
          \STATE $e\leftarrow$ torch.addcmul($-p, a, b$)
           \STATE {\bfseries Return:} $(p,e)$
        \end{algorithmic}
\end{algorithm}
\vspace{-3mm}

The constraint of decreasing magnitude and non-overlapping across $nc$ might be temporarily violated in computation, so MCTensor must be renormalized during computation.  Users can specify the target $nc$ after renormalization, $r_{nc}$, but by default we keep them the same.  The \pythoninline{Renormalize} function is this paper is a variant of the Priest's algorithm~\ref{alg:renormalize}. \cite{priest1991algorithms}.
\vspace{-3mm}
\begin{algorithm}[H]
\caption{\textbf{Renormalize} \cite{priest1992properties}}
\label{alg:renormalize}
\begin{algorithmic}
   \STATE {\bfseries Input:}  $nc$-MCTensor $x$,\space  $r_{nc}$
   \STATE {\bfseries Requires:} $r_{nc} < nc$
   \STATE \textbf{initialize} $s \leftarrow x_0, k \leftarrow 0, t_0 \leftarrow 0$
   
   \FOR{$i=1$ to $nc$}
   \STATE $(s, t_i) \leftarrow \textbf{Two-Sum}(x_i, s)$
   \ENDFOR
   \FOR{$i=0$ to $nc-1$}
   \STATE $(s, e) \leftarrow \textbf{Two-Sum}(s, t_i)$
   \IF{$e\neq 0$}
   \STATE $b_k\leftarrow s$
   \STATE $s\leftarrow e$
   \STATE $k\leftarrow k+1$
   \ENDIF
   \ENDFOR
   \STATE {\bfseries Return:} $(b_0 ,b_1, \cdots ,b_{r_{nc}-1})$
\end{algorithmic}
\end{algorithm}

\vspace{-3mm}

There is a simple and fast implementation of \pythoninline{Renormalize} as \pythoninline{Simple-Renorm}, which extracts all non-zero values from a non-renormalized MCTensor and puts together a new MCTensor.  Note that this \pythoninline{Simple-Renorm} does not have the same guarantee as \pythoninline{Renormalize}. Therefore in our \pythoninline{Mult-MCN}, we still use the original version of \pythoninline{Renormalize}.  But for most other operations, we still utilize \pythoninline{Simple-Renorm} for fast computation.
\vspace{-3mm}
\begin{algorithm}[H]
        \caption{\textbf{Simple-Renorm}}
        \label{alg:simpleRenorm}
        \begin{algorithmic}
          \STATE {\bfseries Input:}  $nc$-MCTensor $x$,\space $r_{nc}$
          \STATE {\bfseries Requires:} $r_{nc} < nc$
        \STATE $k \leftarrow 0;\  (b_0, b_1 ,\cdots ,b_{r_{nc}-1}) \leftarrow (0,0, \cdots, 0)$
          \FOR{$i=0$ to $r_{nc}-1$}
            \IF{$x_i \neq 0$} 
            \STATE $b_{k}\leftarrow x_i$
            \STATE ${k}\leftarrow k + 1$
            \ENDIF
          \ENDFOR
           \STATE {\bfseries Return:} $(b_0, b_1 ,\cdots ,b_{r_{nc}-1})$
        \end{algorithmic}
\end{algorithm}
\vspace{-3mm}

\newpage
Algorithm~\ref{alg:scalingN} describes the multiplication of a MCTensor with a PyTorch Tensor.  In our implementation, the user can specify whether the algorithm can return an expanded results with $nc+1$, or a MCTensor with same $nc$.
\vspace{-3mm}
\begin{algorithm}[H]
   \caption{\textbf{ScalingN}, modified from \cite{shewchuk1997adaptive}}
   \label{alg:scalingN}
    \begin{algorithmic}
    \STATE {\bfseries Input:}  $nc$-MCTensor $x$,\space  PyTorch Tensor $v$, \space $expand$
   \STATE \textbf{initialize} $e \leftarrow 0$
   \FOR{$i=0$ to $nc-1$}
   \STATE $(h_p, e_1) \leftarrow \textbf{Two-Prod}(x_{i}, v)$
   \STATE $(h_i, e_2) \leftarrow \textbf{Two-Sum}(h_{p}, e)$
   \STATE $e \leftarrow  \textbf{fl}(e_1+e_2) $
   \ENDFOR
    \STATE $h  \leftarrow (h_0, \cdots,h_{nc-1},e)$
    \IF{$expand$ is True}
     \STATE{\bfseries Return:} $\textbf{Simple-Renorm}(h, nc+1)$ 
     \ELSE 
     \STATE{\bfseries Return:}
     $\textbf{Simple-Renorm}(h, nc)$
    \ENDIF
\end{algorithmic}
\end{algorithm}
 
\vspace{-3mm}

\pythoninline{Add-MCN, Div-MCN, Mult-MCN} are operators for addition, division, and multiplication of two $nc$-MCTensors. Here we have two versions of multiplication, \pythoninline{Mult-MCN} and \pythoninline{Mult-MCN-Slow}. Algorithm~\ref{alg:multmcn} is implemented by taking the inverse of the second MCTensor first, and then the first MCTensor is divided by the second MCTensor's inverse. This division would give the result of multiplication. Algorithm~\ref{alg:multmcn_slow} follows the same pattern of our definition of \pythoninline{Div-MCN} and provides better error bounds, but it is rarely used as the computational cost is too high.
\vspace{-3mm}
\begin{algorithm}[H]

  \caption{\textbf{Add-MCN}, modified from \cite{shewchuk1997adaptive}}
  \label{alg:addmcn}
    \begin{algorithmic}
      \STATE {\bfseries Input:} $nc$-MCTensor $x, y$
      \STATE \textbf{initialize:} $e \leftarrow 0$
      \FOR{$i=0$ to $nc-1$}
      \STATE $(h_p, e_1) \leftarrow \textbf{Two-Sum}(x_i, y_i)$
      \STATE $(h_i, e_2) \leftarrow \textbf{Two-Sum}(h_{p}, e)$
      \STATE $e \leftarrow  \textbf{fl}(e_1+e_2) $
      \ENDFOR
      \STATE $h  \leftarrow (h_0, \cdots,h_{nc-1},e)$
      \STATE {\bfseries Return:} \textbf{Simple-Renorm}$(h, nc)$
    \end{algorithmic}
\end{algorithm}
\vspace{-3mm}
\begin{algorithm}[H]
\caption{\textbf{Div-MCN}, modified from \cite{shewchuk1997adaptive}}
\label{alg:divmcn}
    \begin{algorithmic}
        \STATE {\bfseries Input:} $nc$-MCTensor $x, y$
        \STATE\bfseries{initialize:} $q\leftarrow \textbf{fl}(x_0/y_0), \space  h_0\leftarrow q$
        \FOR{$i=1$ to $nc$}
        \STATE $r\leftarrow \textbf{Add-MCN}(x, -\textbf{ScalingN}(y, q, \text{False}))$
        \STATE $x\leftarrow r$
        \STATE $q\leftarrow \textbf{fl}(x_0 / y_0)$
        \STATE $h_i\leftarrow q$
        \ENDFOR
        \STATE $h  \leftarrow (h_0, h_{1},\cdots, h_{nc})$
        \STATE {\bfseries Return:} \textbf{Simple-Renorm}$(h, nc)$
    \end{algorithmic}
\end{algorithm}
\vspace{-3mm}
\begin{algorithm}[H]
\caption{\textbf{Mult-MCN} }
\label{alg:multmcn}
    \begin{algorithmic}
        \STATE {\bfseries Input:} $nc$-MCTensor $x, y$
        \STATE\bfseries{initialize:} $z_0 \leftarrow 1, \space z_1=\cdots=z_{nc-1}\leftarrow 0$
         \STATE $z \leftarrow (z_0,z_1,\cdots ,z_{nc-1})$
        \STATE $y^{-1} \leftarrow  \textbf{Renormalize}(\textbf{Div-MCN}(z, y))$
        \STATE $h \leftarrow \textbf{Renormalize}(\textbf{Div-MCN}(x, y^{-1}))$
        \STATE {\bfseries Return:}  $h$
    \end{algorithmic}
\end{algorithm}
\vspace{-3mm}
\begin{algorithm}[H]
\caption{\textbf{Mult-MCN-Slow} }
\label{alg:multmcn_slow}
    \begin{algorithmic}
        \STATE {\bfseries Input:} $nc$-MCTensor $x, y$
        \STATE\bfseries{initialize:} $p\leftarrow \textbf{fl}(x_0\cdot y_0), \space  h_0\leftarrow p$
        \FOR{$i=1$ to $nc$}
        \STATE $e\leftarrow \textbf{Add-MCN}(x, -\textbf{DivN}(p, y))$
        \STATE $x\leftarrow e$
        \STATE $p\leftarrow \textbf{fl}(x_0 \cdot y_0)$
        \STATE $h_i\leftarrow p$
        \ENDFOR
        \STATE $h  \leftarrow (h_0, h_{1},\cdots, h_{nc})$
        \STATE {\bfseries Return:} \textbf{Simple-Renorm}$(h, nc)$
    \end{algorithmic}
\end{algorithm}
\vspace{-3mm}

Algorithm~\ref{alg:divN}, \pythoninline{DivN} takes input of a PyTorch Tensor and a MCTensor and computes the \pythoninline{Div-MCN} results by appending zero-value components to a PyTorch Tensor and making it MCTensor.
\vspace{-3mm}
\begin{algorithm}[H]
\caption{\textbf{DivN} }
\label{alg:divN}
    \begin{algorithmic}
        \STATE {\bfseries Input:} PyTorch Tensor $x_0$, $nc$-MCTensor $y$, 
        \STATE\bfseries{initialize:} $  x_1=\cdots=x_{nc-1}\leftarrow 0$
        \STATE $x\leftarrow (x_0,x_1,\cdots,x_{nc-1})$
        
        \STATE {\bfseries Return:} \textbf{Div-MCN}$(x, y)$
    \end{algorithmic}
\end{algorithm}

\newpage
The following Algorithm~\ref{alg:expn} describes the exponential function for a MCTensor.
\vspace{-3mm}
\begin{algorithm}[H]
   \caption{\textbf{Exp-MCN}}
   \label{alg:expn}
\begin{algorithmic}
   \STATE {\bfseries Input:}  $nc$-MCTensor $x$ 
   \STATE \textbf{initialize} $ h \leftarrow \text{exp}(x_0)$
   \FOR{$i=1$ to $nc-1$}
    \STATE $h \leftarrow \textbf{ScalingN}(h, \text{exp}(h_i), \text{True})$
   \ENDFOR
   \STATE {\bfseries Return:} $h$ 
\end{algorithmic}
\end{algorithm}

The following Algorithm~\ref{alg:SquareN} describes the square of a MCTensor.
\vspace{-3mm}
\begin{algorithm}[H]
   \caption{\textbf{Square-MCN}}
   \label{alg:SquareN}
\begin{algorithmic}
   \STATE {\bfseries Input:}  $nc$-MCTensor $x$ 
   \STATE \textbf{initialize} $ h_0 \leftarrow \textbf{fl}(2^{x_0}), \space h_1=\cdots= h_{nc-1} \leftarrow 0$
   
   \STATE  $ h \leftarrow (h_0,h_1,\cdots ,h_{nc-1})$
    
    \STATE $h \leftarrow \textbf{Grow-ExpN}(h,  \textbf{fl}(2\cdot x_0\cdot x_1))$
    
   \STATE {\bfseries Return:} $h$ 
\end{algorithmic}
\end{algorithm}
\vspace{-3mm}

\subsubsection{Matrix Operators}
Based on the dimensions of input MCTensor and PyTorch Tensor, \pythoninline{Dot-MCN, MV-MCN, MM-MCN, BMM-MCN and 4DMM-MCN} are implemented for calculating the matrix-level multiplication results.  All operations are identical with the PyTorch implementations.  
\vspace{-3mm}
\begin{algorithm}[H]
   \caption{\textbf{Dot-MCN} }
   \label{alg:dotmcn}
\begin{algorithmic}
    \STATE {\bfseries Input:}  $nc$-MCTensor $x$,\space  PyTorch Tensor $v$ 
    \STATE {\bfseries Requires:} $x$ and $v$ both  1D array
    \STATE $h  \leftarrow \textbf{ScalingN}(x, v, \text{False})$
    \STATE $h_{tensor} = h_0 + h_1 + \cdots + h_{nc-1}$
    
    \STATE {\bfseries Return:}  $h_{tensor}$
    
\end{algorithmic}
\end{algorithm}
 
\vspace{-3mm}
\begin{algorithm}[H]
   \caption{\textbf{MV-MCN} }
   \label{alg:mvmcn}
\begin{algorithmic}
    \STATE {\bfseries Input:}  $nc$-MCTensor $x$,\space  PyTorch Tensor $v$ 
    \STATE {\bfseries Requires:} $x$ is 2D matrix of size $(n\times m)$ and $v$ is 1D array of size $m$ 
    \STATE $scaled  \leftarrow \textbf{ScalingN}(x, v, \text{False})$
    \STATE $h  \leftarrow scaled[...,0]$
    \FOR{$i=1$ to $m-1$}
        \STATE $h \leftarrow \textbf{Add-MCN}(h, scaled[...,i])$
        \ENDFOR
    \STATE {\bfseries Return:}  $h $
    
\end{algorithmic}
\end{algorithm}
 
\vspace{-3mm}
\begin{algorithm}[H]
   \caption{\textbf{MM-MCN} }
   \label{alg:mmmcn}
\begin{algorithmic}
    \STATE {\bfseries Input:}  $nc$-MCTensor $x$,\space  PyTorch Tensor $v$ 
    \STATE {\bfseries Requires:} $x$ is 2D matrix of size $(n\times m)$ and $v$ is 2D Matrix of size $(m \times p)$ 
     \STATE $x \leftarrow x.\textbf{unsqueeze}(-1)$
     \STATE $v \leftarrow v.\textbf{transpose}(-1, -2)$
    \STATE $scaled  \leftarrow \textbf{ScalingN}(x, v, \text{False})$
    \STATE $h  \leftarrow scaled[...,0]$
    \FOR{$i=1$ to $m-1$}
        \STATE $h \leftarrow \textbf{Add-MCN}(h, scaled[...,i])$
        \ENDFOR
    \STATE {\bfseries Return:}  $h $
    
\end{algorithmic}
\end{algorithm}
 
\vspace{-3mm}
\begin{algorithm}[H]
   \caption{\textbf{BMM-MCN} }
   \label{alg:bmmmcn}
\begin{algorithmic}
    \STATE {\bfseries Input:}  $nc$-MCTensor $x$,\space  PyTorch Tensor $v$ 
    \STATE {\bfseries Requires:} $x$ is 3D matrix of size $(b \times n\times m)$ and $v$ is 3D Matrix of size $(b \times m \times p)$ 
     \STATE $x \leftarrow x.\textbf{unsqueeze}(-1)$
     \STATE $v \leftarrow v.\textbf{unsqueeze}(1).\textbf{transpose}(-1, -2)$
    \STATE $scaled  \leftarrow \textbf{ScalingN}(x, v, \text{False})$
    \STATE $h  \leftarrow scaled[...,0]$
    \FOR{$i=1$ to $m-1$}
        \STATE $h \leftarrow \textbf{Add-MCN}(h, scaled[...,i])$
        \ENDFOR
    \STATE {\bfseries Return:}  $h $
    
\end{algorithmic}
\end{algorithm}
 
\vspace{-3mm}
\begin{algorithm}[H]
   \caption{\textbf{4DMM-MCN} }
   \label{alg:4dmmmcn}
\begin{algorithmic}
    \STATE {\bfseries Input:}  $nc$-MCTensor $x$,\space  PyTorch Tensor $v$ 
    \STATE {\bfseries Requires:} $x$ is 4D matrix of size $(a\times b \times n\times m)$ and $v$ is 4D Matrix of size $(c\times d \times m \times p)$; two sizes can be broadcasted in \textbf{torch.matmul} 
     \STATE $x \leftarrow x.\textbf{unsqueeze}(-1)$
     \STATE $v \leftarrow v.\textbf{unsqueeze}(2).\textbf{transpose}(-1, -2)$
    \STATE $scaled  \leftarrow \textbf{ScalingN}(x, v, \text{False})$
    \STATE $h  \leftarrow scaled[...,0]$
    \FOR{$i=1$ to $m-1$}
        \STATE $h \leftarrow \textbf{Add-MCN}(h, scaled[...,i])$
        \ENDFOR
    \STATE {\bfseries Return:}  $h $
    
\end{algorithmic}
\end{algorithm}
 
\vspace{-3mm}
The following Algorithm~\ref{alg:addmmmcn} computes the matrix multiplication of a $nc$-MCTensor with a PyTorch Tensor first, then times a constant $\alpha$.  Then the product of another $nc$-MCTensor times a constant $\beta$ is added to the former multiplication result. 
\begin{algorithm}[H]
   \caption{\textbf{AddMM-MCN} }
   \label{alg:addmmmcn}
\begin{algorithmic}
    \STATE {\bfseries Input:}  $nc$-MCTensor $x$, $nc$-MCTensor $y$, PyTorch Tensor $v$, $\beta$, $\alpha$
    \STATE {\bfseries Requires:} $x$ is 2D matrix of size $(n\times m)$, and $v$ is 2D Matrix of size $(m \times p)$; $y$ is 2D matrix of size $(n\times m)$ or $y$ is 1D array of size $m$
    \STATE $h \leftarrow \textbf{ScalingN}( \textbf{MM-MCN}(x, v), \alpha)$
     \STATE $bias \leftarrow  \textbf{ScalingN}(y, \beta)$
     \STATE $h \leftarrow  \textbf{Add-MCN}(h, bias)$ 
    \STATE {\bfseries Return:}  $h $
    
\end{algorithmic}
\end{algorithm}
 
\vspace{-3mm}
\newpage
This Algorithm~\ref{alg:matmulmcn}, \pythoninline{Matmul-MCN} is the central function for handling all matrix level multiplications of one $nc$-MCTensor and one PyTorch Tensor.
\begin{algorithm}[H]
   \caption{\textbf{Matmul-MCN} }
   \label{alg:matmulmcn}
\begin{algorithmic}
    \STATE {\bfseries Input:}  $nc$-MCTensor $x$,  PyTorch Tensor $y$ 
    \STATE $x_d, y_d \leftarrow x.\textbf{dim()}, y.\textbf{dim()}$
     \IF{$x_d = 1$ and $y_d = 1$}
     \STATE {\bfseries Return:}  $\textbf{Dot-MCN}(x,y) $
     \ELSIF{$x_d = 2$ and $y_d = 2$}
     \STATE {\bfseries Return:}  $\textbf{MM-MCN}(x,y) $
     \ELSIF{$x_d = 2$ and $y_d = 1$}
     \STATE {\bfseries Return:}  $\textbf{MV-MCN}(x,y) $
     \ELSIF{$x_d > 2$ and $y_d = 1$}
     \STATE {\bfseries Return:}  $\textbf{ScalingN}(x,y) $
     \ELSIF{$x_d = y_d $ and $x_d = 3$}
     \STATE {\bfseries Return:}  $\textbf{BMM-MCN}(x,y) $
     \ELSIF{$x_d = y_d $ and $x_d = 4$}
     \STATE {\bfseries Return:}  $\textbf{4DMM-MCN}(x,y) $
     \ENDIF

\end{algorithmic}
\end{algorithm}
 
\vspace{-3mm}

\onecolumn



\subsection{MCF Basic and Matrix Operators } \label{mm}
We run on a AMD Ryzen 7 5800X CPU with 64 GB memory. 

For all the basic operators, we repeat PyTorch addition for 7e3 runs and 1-, 2-, 3-MCTensor for 7 runs.

For vector-vector (\pythoninline{torch.dot}/\textbf{Dot-MCN}) product, we repeat PyTorch addition for 7e6 runs and 1-, 2-, 3-MCTensor for 7e3 runs. For matrix-vector (\pythoninline{torch.mv}/\textbf{MV-MCN}) product and batched matrix-matrix (\pythoninline{torch.bmm}/\textbf{BMM-MCN}) product, we repeat PyTorch addition for 7e3 runs and 1-, 2-, 3-MCTensor for 7 runs. For matrix-matrix (\pythoninline{torch.matmul}/\textbf{Matmul-MCN}) product and bias-matrix-matrix (\pythoninline{torch.addmm}/\textbf{AddMM-MCN}) addition-product, we repeat PyTorch addition for 7e4 runs and 1-, 2-, 3-MCTensor for 7 runs. 
 
The timing for both matrix and basic operators are given in Table \ref{tab:basicop} and Table \ref{tab:matrixop_full}.
\begin{table*}[h]\centering
    
    \scalebox{0.85}{\small\begin{tabular}{lccccr}
    \toprule
     Operators & Inputs sizes & FloatTensor & 1-MCTensor & 2-MCTensor & 3-MCTensor  \\
    \midrule
    \textbf{Add-MCN} & $(1000\times 1000)$  & $497 \ \us \pm 6.77\ \us$
                                         & $26.7 \ \ms \pm 486 \ \us$
                                         & $44.7 \ \ms \pm 379 \ \us$
                                         & $64.4 \ \ms \pm 385 \ \us$\\                      
    \textbf{ScalingN} & $(1000\times 1000)$    & $ 490  \ \us \pm 9.69 \ \us$
                                             & $ 33.7  \ \ms \pm 402  \ \us$
                                             & $ 57.5 \ \ms \pm 842  \ \us$
                                             & $ 84.7 \ \ms \pm 1.78 \ \ms$ \\
    \textbf{Mult-MCN} & $(1000\times 1000)$  & $ 514   \ \us \pm 15.2\ \us$
                                             & $ 218   \ \ms \pm 4.03 \ \ms$
                                             & $ 667  \ \ms \pm 11.6 \ \ms$
                                             & $ 1.4 \ s \pm 21.6 \ \ms$ \\
    \textbf{Div-MCN} & $(1000\times 1000)$ & $ 510    \ \us \pm 10.6 \ \us$
                                             & $ 80.3   \ \ms \pm 770 \ \us$
                                             & $ 243  \ \ms \pm 3.24 \ \ms$
                                             & $ 498  \ \ms \pm 8.26 \ \ms$ \\ 
    \bottomrule
    \end{tabular}}
    \caption{\centering  MCTensor Basic Operators Running Time (mean $\pm$ sd) }
    \label{tab:basicop}
    \vspace{-1em}
\end{table*}

\begin{table*}[h]
 \centering
   
    \scalebox{0.8}{\small\begin{tabular}{lccccl} 
    \toprule
     Operators & Inputs sizes & FloatTensor & 1-MCTensor & 2-MCTensor & 3-MCTensor  \\
    \midrule
    \textbf{Dot-MCN} & $5000, 5000$ 
    & $ 1.61\ \us\pm3.29\ \ns$ 
    & $ 442\ \us\pm5.61\ \us$
    & $ 656\ \us\pm1.16\ \us$
    & $ 858\ \us\pm12.2\ \us$\\                
\textbf{MV-MCN} & $(5000\times 500),500$    
& $ 157\ \us \pm4.32\us$
& $ 320\ \ms \pm5.78\ms$
& $ 460\ \ms \pm10.7\ms$
& $ 580\ \ms \pm12.1\ms$ \\
    \textbf{Matmul-MCN} 
    & ( 500$\times$ 200), ( 200$\times$ 50)    
    & $ 97.3\ \us \pm1.1\ \us$
    & $ 495\ \ms \pm10.8\ \ms$
    & $ 735\ \ms \pm21.7 \ \ms$
    & $ 934\ \ms \pm28 \ \ms$ \\
\textbf{AddMM-MCN} & 100, ( 500$\times$ 200), ( 200$\times$ 100),   
    & $ 153  \ \us \pm 869 \ \ns$
    & $ 750  \ \ms \pm 21.9 \ \ms$
    & $ 1.12  \ \text{s} \pm 28.4 \ \ms$
    & $ 1.44   \ \text{s} \pm 49.1 \ \ms$ \\  
\textbf{BMM-MCN} & (16 $\times$ 500$\times$ 200), (16 $\times$200$\times$ 50)  
                                            & $ 1.5    \ \ms \pm 9.32 \ \us$
                                             & $ 4.84   \ \text{s} \pm 43.1 \ \ms$
                                             & $ 8.03   \ \text{s} \pm 68.4 \ \ms$
                                             & $ 11.2   \ \text{s} \pm 74.1 \ \ms$ \\
                                 
    \bottomrule

    \end{tabular}}
     \caption{\centering  MCTensor Matrix Operators Running Time (mean $\pm$ sd) }
             \label{tab:matrixop_full}
    \vspace{-1em}
\end{table*}

\subsection{Linear Regression Results} \label{LR}
Table \ref{tab:linreg} describes the final training loss of the linear regression task in section \ref{subsec:error}.
\begin{table}[ht] 
     \centering
    \scalebox{0.85}{\small
    \begin{tabular}{lccccr}
    \toprule
     \textbf{Model} & \textbf{Train Loss}  \\
    \midrule
    PyTorch float16 & 1.99e-4 \\ 
    PyTorch float32 & 2.64e-12 \\ 
    PyTorch float64 & 8.02e-18 \\ 
    MCTensor float16, nc = 1 & 1.80e-4 \\ 
    MCTensor float32, nc = 2 & 1.95e-7 \\ 
    MCTensor float64, nc = 3 & 1.95e-7 \\ 
    \bottomrule
    \end{tabular}
    }
    \caption{\centering Final Training Loss Results of Linear Regression Task}
        \label{tab:linreg}
\end{table}

\newpage

\subsection{Logistic Regression}\label{logreg}
We apply logistic regression on a synthetic dataset and a breast cancer dataset. The synthetic dataset consists of 1,000 data points, where each data point contains two features, while the breast cancer dataset consists of 569 data points and each data point contains 30 numeric features. A single \pythoninline{MCLinear} layer with float16 and $nc$ between 1, 2 and 3 are used for logistic regression with Binary Cross Entropy loss.

$$ \mathcal{L}(W) = \pythoninline{BCELoss}(y, \text{sigmoid}(XW^T)) $$

For both datasets, we randomly split 80\% of the data for training and 20\% of the data for testing. As both datasets are small in scale, we run them in full batch with MC-SGD and SGD. For the synthetic dataset, we set learning rate to be 3e-3 and run for 4000 epochs. For the breast cancer dataset, we set learning rate to be 1e-4 and momentum as 0.9, and run for 3000 epochs. Below is the results for both dataset and figures for MCLinear results on breast cancer dataset with MC-SGD optimizer. As can be seen, even with as few as 2 components, MCTensor in float16 can match with the training loss of float32 Tensor. 
\begin{figure*}[ht] \vspace{-1em}
\centering
\begin{subfigure}
\centering 
\includegraphics[width=0.4\textwidth]{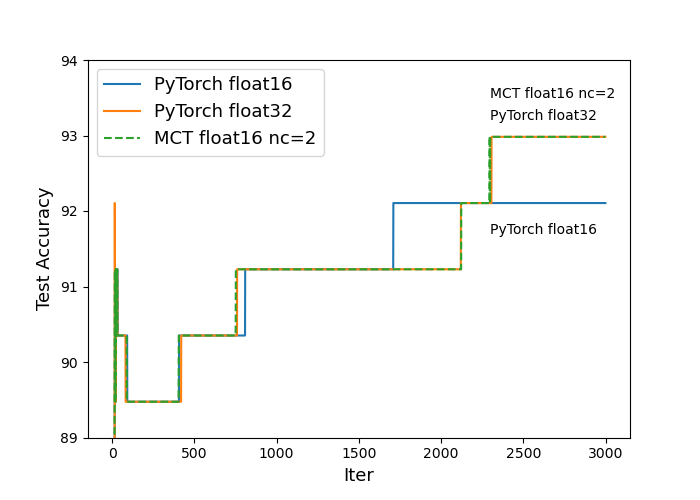}  
\end{subfigure} \vspace{-1em}
\caption{\centering Testing Accuracy for MCTensor and PyTorch Logistic Regression on the Breast Cancer Dataset} 
\label{fig:logistic_cancer}
\end{figure*}
 
\begin{table*}[h]
    \centering
    \scalebox{1}{\small\begin{tabular}{lccccr}
\toprule
 \textbf{Model} & \textbf{Training Loss} &    \textbf{Testing Accuracy (\%)} \\
\midrule
Tensor float16 &  0.1940  &  100 \\ 
 {Tensor float32} & 0.1042  &  100 \\ 
 {Tensor float64} & 0.1042  &  100 \\ 
1-MCTensor float16  & 0.1941  &  100 \\ 
\textbf{2-MCTensor float16}  & \textbf{0.1041}  &  100 \\ 
\textbf{3-MCTensor float16}  & \textbf{0.1041}  &  100 \\ 
\bottomrule
\end{tabular}}


 \vspace{-1em}
      \captionof{table}{ \centering Final Training and Testing Results for Logistic Regression on Synthetic Dataset}
\end{table*}
 
\begin{table*}[h]
    \centering
    \scalebox{1}{\small\begin{tabular}{lccccr}
\toprule
 \textbf{Model} & \textbf{Training Loss} & \textbf{Testing Accuracy (\%)} \\
\midrule
Tensor float16 &  0.1944  & 92.11 \\ 
Tensor float32 & 0.1528  & \textbf{92.98} \\ 
Tensor float64 & 0.1528  & \textbf{92.98} \\ 
1-MCTensor float16  & 0.1944  & 92.11 \\ 
\textbf{2-MCTensor float16}  & \textbf{0.1527} & \textbf{92.98} \\ 
\textbf{3-MCTensor float16}  & \textbf{0.1527}  & \textbf{92.98} \\ 
\bottomrule
\end{tabular}}


    \vspace{-1em}
      \captionof{table}{ \centering Final Training and Testing Results for Logistic Regression on Breast Cancer Dataset}
\end{table*}
 \newpage
 \subsection{MCTensor Deep Learning models} 

As a demonstration for the ease of using MCTensor to build a MCTensor deep learning model, we provide an example code for MCTensor MLP (MC-MLP) model for the multi-class classification tasks. Essentially, the only differences visible to the users are the need to set the number of components, and the need to convert MCTensor output to Tensor approximation between different layers. The need for Tensor approximation is explained below. \par

\begin{figure}[ht]
\centering
\pythonexternal{code/MLP_toy.py}
\caption{MC-MLP Code Example}
\label{fig:MLP}
\end{figure}

In MCModule, all network layers take Tensor as inputs, keep MCTensor as their weights, and through MCTensor matrix operations, produces MCTensor as output. If the MCTensor outputs need to be taken as input for the next MCModule layer, there need to be new layers that will perform multiplication on two MCTensor matrices. As can be seen from Table \ref{tab:basicop}, \textbf{ScalingN}, the multiplication between a MCTensor and a PyTorch Tensor, is more than a hundred times faster than \textbf{Mult-MCN}, the multiplication between two MCTensor. To avoid this forbidden cost of computation, we convert the unevaluated sums into an evaluated sum as shown above (\pythoninline{x.tensor.sum(-1)}). Although there might some losses of precision due to summation, the loss is marginal and we trade it for an orders of magnitude smaller execution time.

\subsection{MC-MLP Experiment Details}\label{MC-MLP}

Using MC-MLP, we perform multi-class classification task on the Reduced MNIST dataset and binary classification task on the Breast Cancer dataset. The training details for both tasks are shown in Table \ref{tab:mlp-b-hyper} and Table \ref{tab:mlp-m-hyper}.  
\begin{table}[ht] 
 \centering
    \scalebox{0.85}{
        \small\begin{tabular}{lcr}
        \toprule
         \textbf{Parameter} & \textbf{Value}  \\
        \midrule
        \textbf{MLP first hidden layer dim} & 150 \\
        \textbf{MLP second hidden layer dim} & 150 \\
        \textbf{Batch size} & full batch (569) \\
        \textbf{Optimizer} & MC-SGD (GD with full batch) \\
        \textbf{Learning rate} & 6e-3\\
        \textbf{Epoch} & 1000 \\
        \bottomrule
        \end{tabular}
    }
    \caption{\centering Training Details for the Breast Cancer dataset}
        \label{tab:mlp-b-hyper}
\end{table}

\begin{table}[ht] 
 \centering
    \scalebox{0.85}{
        \small\begin{tabular}{lcr}
        \toprule
         \textbf{Parameter} & \textbf{Value}  \\
        \midrule
        \textbf{MLP first hidden layer dim} & 50 \\
        \textbf{MLP second hidden layer dim} & 50 \\
        \textbf{Batch size} & 128 \\
        \textbf{Optimizer} & MC-SGD (SGD) \\
        \textbf{Learning rate} & 2e-3\\
        \textbf{Momentum} & 0.8  \\
        \textbf{Epoch} & 100\\
        \bottomrule
        \end{tabular}
    }
    \caption{Training Details for the Reduced MNIST Dataset}
        \label{tab:mlp-m-hyper}
\end{table}

\begin{figure}[ht]\vspace{-0.5em}
\centering
\includegraphics[width=0.4\textwidth]{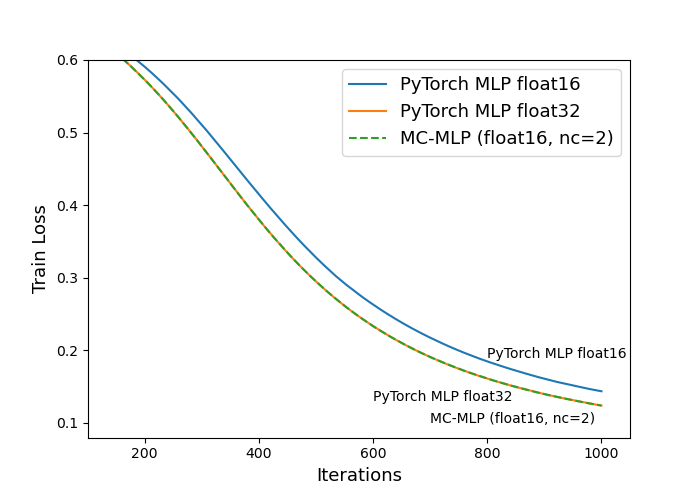}
\caption{\centering Training Loss Curves for MLP on Breast Cancer Dataset}
\label{fig:mlp_c_sgd}
\end{figure}

\newpage
Figure \ref{fig:mlp_c_sgd} shows the training loss curves for MC-MLP on breast cancer dataset with MC-SGD optimizer. 

\begin{figure}[!ht]
\centering
\includegraphics[width=0.5\textwidth]{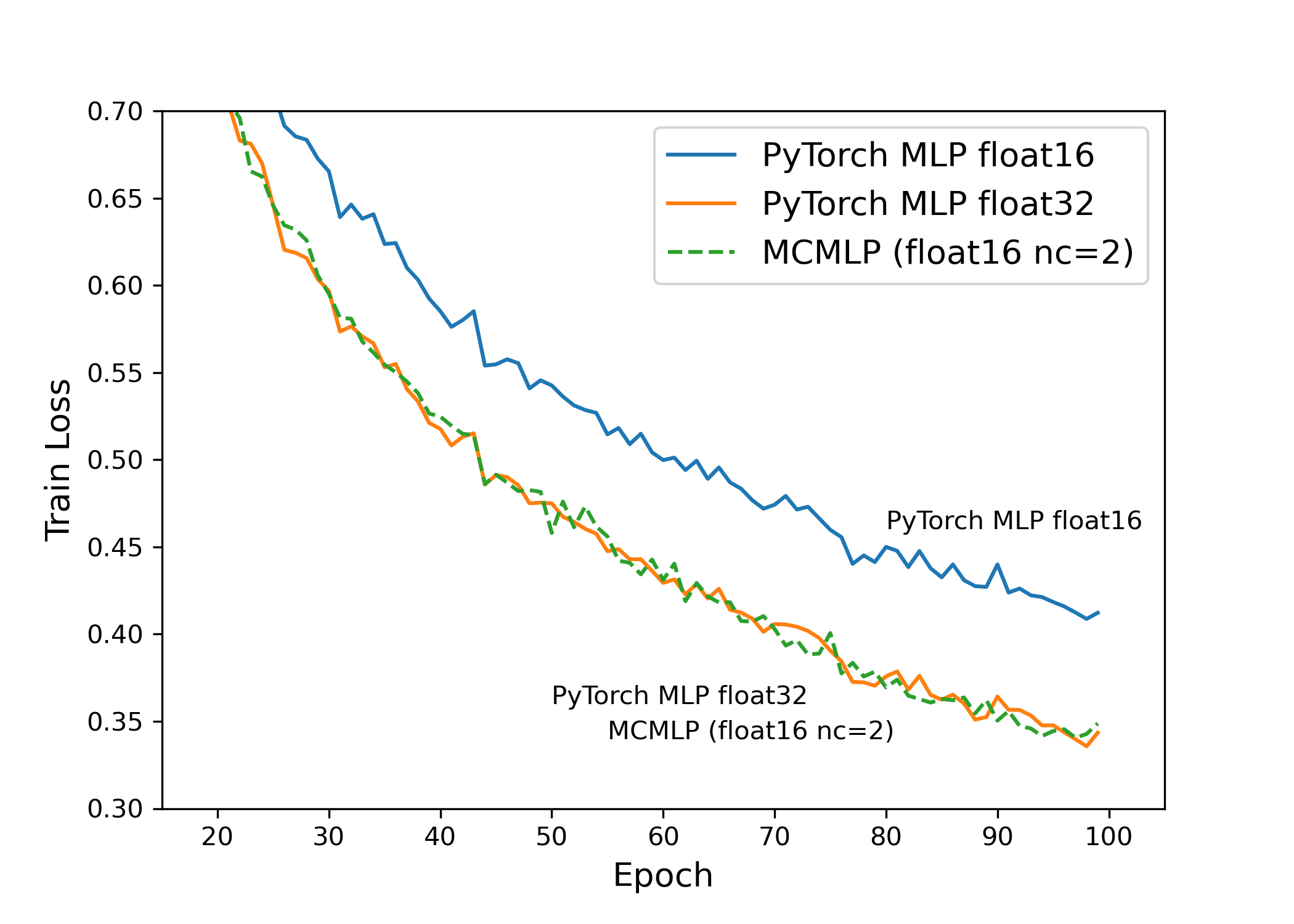}
\vspace{-1em}
\caption{Training Loss Curves for MLP on Reduced MNIST Dataset}
\label{fig:mlp_m_sgd}
\end{figure}

Figure \ref{fig:mlp_m_sgd} shows the training loss curves for MC-MLP on reduced MNIST dataset with MC-SGD optimizer. The hyperparameters used in training are shown in Table \ref{tab:mlp-m-hyper}, and the final results for testing accuracy is shown in Table \ref{tab:mlp-m}.


\begin{table}[ht]
 \centering
    \scalebox{0.85}{
        \small\begin{tabular}{lccr}
        \toprule
         Model & Training Loss & Testing accuracy   \\
        \midrule
        \textbf{MCMLP (f16 nc=1)} & 0.424 & 91.40 \\
        \textbf{MCMLP (f16 nc=2)} & 0.349 & \textbf{92.03}\\
        \textbf{MCMLP (f16 nc=3)} & 0.349 & 91.98\\
        \textbf{PyTorch MLP (f16)} & 0.412 & 91.40 \\
        \textbf{PyTorch MLP (f32)} & \textbf{0.343} & 92.00 \\
        \textbf{PyTorch MLP (f64)} & \textbf{0.343} & 92.00\\
        \bottomrule
        \end{tabular}
    }
    \caption{\centering MC-MLP Models on Reduced MNIST Dataset with MCSGD}
    \label{tab:mlp-m}
\end{table}



\end{document}